\documentclass{ieeeoj}
\usepackage{cite}
\usepackage{amsmath,amssymb,amsfonts}
\usepackage{graphicx,color}
\usepackage{textcomp}
\def\BibTeX{{\rm B\kern-.05em{\sc i\kern-.025em b}\kern-.08em
    T\kern-.1667em\lower.7ex\hbox{E}\kern-.125emX}}
\AtBeginDocument{\definecolor{ojcolor}{cmyk}{0.93,0.59,0.15,0.02}}
\def\OJlogo{\vspace{-14pt}\includegraphics[height=28pt]{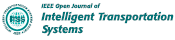}}

\usepackage{booktabs}
\usepackage{makecell}
\usepackage{multirow}

\graphicspath{{figures/}{./}}
\DeclareGraphicsExtensions{.pdf,.jpeg,.png}

\usepackage[caption=false,font=footnotesize]{subfig}

\usepackage{url}
\usepackage{nameref}
\makeatletter
\AtBeginDocument{%
  \let\ieeeoj@sect\@sect
  \def\@sect#1#2#3#4#5#6[#7]#8{%
    \def\@currentlabelname{#8}%
    \ieeeoj@sect{#1}{#2}{#3}{#4}{#5}{#6}[{#7}]{#8}%
  }%
}
\makeatother
\usepackage{dblfloatfix}
\usepackage{placeins}



\begin{document}

\title{A Sparsity Analysis of \\ Traffic Forecasting Architectures: \\ How Much Missing Data is Too Much?}

\author{Mohamed Hamad\authorrefmark{1}, Bilal Mutahir\authorrefmark{1}, Mohamed Mabrok\authorrefmark{2}, Nizar Zorba\authorrefmark{1}, Hamid Saeedi\authorrefmark{3}, and Halim Yanikomeroglu\authorrefmark{4}}

\affil{Department of Electrical Engineering, Qatar University, Doha, Qatar}
\affil{Department of Mathematics and Statistics, Qatar University, Doha, Qatar}
\affil{College of Engineering and Technology, University of Doha for Science and Technology, Doha, Qatar}
\affil{Department of Systems and Computer Engineering, Carleton University, Ontario, Canada}

\corresp{CORRESPONDING AUTHOR: Mohamed Mabrok (e-mail: m.a.mabrok@qu.edu.qa).}

\markboth{IEEE Open Journal of Intelligent Transportation Systems}{Hamad \MakeLowercase{\textit{et al.}}: Sparsity Analysis of Traffic Forecasting Architectures}

\begin{abstract}
Traffic forecasting models are typically evaluated under ideal conditions with complete data. This evaluation does not reflect the frequent sensor failures and communication outages common in real-world deployments. This paper presents a comprehensive analysis of multivariate traffic prediction models to evaluate their performance under varying levels of sensor data loss during inference. We benchmark several architectural paradigms, including graph neural networks, attention-based Transformers, and selective state-space models, across sparsity levels ranging from zero to complete sensor loss, using two widely used multi-channel traffic benchmarks. To ensure consistent comparison across traffic variables, we use normalized metrics and define a usability threshold to identify when predictions become unreliable. Our results show a clear trade-off between baseline accuracy and robustness: models with the highest accuracy under full-data conditions often degrade most sharply as data becomes sparse, while less accurate models maintain more stable performance. These findings show that model selection should be based on the expected reliability of the sensor network rather than peak accuracy alone. This work provides a practical framework for assessing traffic forecasting models and highlights the importance of robustness in real-world applications.
\end{abstract}


% IEEE notice for this author-submitted preprint. Overlay the space above the
% title so the body, figures, references, and pagination remain unchanged.
\AddToHookNext{shipout/foreground}{%
  \put(\dimexpr64bp-\hoffset\relax,\dimexpr-48bp+\voffset\relax){%
    \makebox[500bp][c]{\normalfont\fontsize{9}{10}\selectfont
      This work has been submitted to the IEEE for possible publication.}}%
  \put(\dimexpr64bp-\hoffset\relax,\dimexpr-58bp+\voffset\relax){%
    \makebox[500bp][c]{\normalfont\fontsize{9}{10}\selectfont
      Copyright may be transferred without notice, after which this version may no longer be accessible.}}%
}

\maketitle

\section{Introduction}
\label{sec:introduction}
\IEEEPARstart{T}{raffic} forecasting plays a crucial role in modern transportation management systems, enabling authorities to optimize traffic flow, reduce congestion, and improve travel times through accurate prediction of future traffic conditions \cite{11042911}. 
Effective forecasting requires reliable measurements of key traffic variables (speed, flow, and occupancy) collected from sensor networks deployed across transportation infrastructure. However, real-world sensor networks are inherently fragile and commonly suffer from failures, communication outages, or maintenance, leading to chronic data scarcity \cite{huang2025ojits_missing}.
This missing data can be simulated as random sensor sparsity, where portions of the sensor network become unavailable at inference time, creating substantial challenges for accurate prediction.

Despite the prevalence of data sparsity in real-world deployments, traffic forecasting models are predominantly benchmarked under ideal, full-data conditions in the papers that introduce them \cite{shao2024stmamba, liu2023staeformer, wang2021stid},
where complete sensor measurements are assumed to be available at inference time. This evaluation paradigm, while useful for establishing baseline performance, overlooks the practical reality of real-world deployments. Much of the current work that addresses the sparsity issue does so before training, employing data imputation techniques to fill missing values and complete the dataset to avoid hindering the model's training process \cite{huang2025ojits_missing, its_sparsity_2023}. 
However, in reality, data may still be unavailable at inference time due to the same limitations that cause sparsity in the first place. Therefore, analysis that evaluates model performance under inference-time sparsity, where missing data cannot be imputed from historical patterns, is more relevant for assessing real-world deployment suitability.

The challenge of data sparsity reveals a critical limitation in current evaluation practices. Specifically, simple high baseline accuracy is insufficient for real-world suitability \cite{2008guide}.
A model with the best baseline accuracy could be the most fragile under sparse data conditions, while a model with lower baseline accuracy might exhibit superior robustness. This accuracy-robustness trade-off indicates that model selection must be conditioned on expected data availability, not just peak performance. Understanding how different architectures respond to varying levels of sparsity is therefore essential for making informed deployment decisions.

Despite the importance of robustness evaluation, existing studies face several limitations that prevent a comprehensive understanding of model behavior under sparsity. Many robustness studies are limited to single-channel datasets, focusing exclusively on speed-only or flow-only measurements \cite{ijcai_2024_stdmae, DCST-Mamba_2025, leaf_2025}.
However, comprehensive traffic understanding and forecasting require simultaneous prediction of multiple variables, including speed, flow, and occupancy \cite{Seo2017}. 
A complete evaluation must therefore use multi-channel datasets to capture the full complexity of multivariate traffic prediction. Furthermore, scale differences between traffic variables, such as speed measured in miles per hour versus flow measured in vehicles per interval, render direct error comparisons misleading \cite{mcst-mamba_2025}. 
A rigorous evaluation must employ normalized error metrics to provide comparable and channel-averaged scores across different measurement types.

While architectural comparisons exist in the literature, the specific robustness of the latest paradigms, such as State-Space Models, under sensor dropout remains largely unexplored. Additionally, performance evaluation must move beyond arbitrary error figures and tie them to the specific accuracy requirements of transportation applications. Without this connection, it remains unclear whether a model's performance degradation under sparsity renders it unsuitable for practical deployment.

This paper addresses these gaps by introducing a unified framework for inference-time sparsity benchmarking and presenting a comprehensive multivariate evaluation of traffic forecasting models under that protocol. Our contributions are as follows:
\begin{itemize}
\item \textbf{Comprehensive Multivariate Sparsity Analysis:} We conduct a systematic analysis of model performance degradation across a discrete sparsity grid from $0\%$ to $100\%$. Crucially, this evaluation is performed on multi-channel traffic data (speed, flow, occupancy) using the PEMS08 and PEMS04 datasets, providing a more complete picture of model robustness than single-channel studies.
\item \textbf{Architectural Robustness Benchmark:} We provide an in-depth robustness comparison across diverse architectures, including GNNs, Transformers, and State-Space Models, to gain insights into the effect of architectural choices on model robustness to sparsity.
\item \textbf{Use of Normalized, Relative and Horizon-Dependent Metrics:} We employ z-score normalized vehicular error metrics (vMAE and vRMSE) to ensure comparable error values across different traffic variables. Furthermore, we evaluate the models' performance relative to their own baseline performance to effectively quantify the degradation in performance due to sparsity. Additionally, we conduct a per-horizon analysis across 12 prediction steps, revealing how sparsity sensitivity evolves over the forecasting window.
\item \textbf{Integration of the Usability Breakdown Threshold:} We establish a framework for using industry standards for accuracy of predictions to define the limits of the usability of models. This is done by defining the breakdown point as the sparsity level where error exceeds a specified application-derived threshold. This provides practical, operationally-relevant benchmarks for model selection and deployment.
\end{itemize}

The remainder of this paper is organized as follows: Section~\ref{sec:related_work} reviews related work on traffic forecasting models and sparsity analysis. Section~\ref{sec:methodology} describes our methodology including our sparsity simulation technique, evaluation framework, dataset description, and model selection. Section~\ref{sec:results_and_analysis} presents our experimental results and analysis. Section~\ref{sec:discussion} discusses the findings and implications. Finally, Section~\ref{sec:conclusion} concludes the paper.

\section{Related Work}
\label{sec:related_work}

\subsection{Traffic Forecasting Datasets}
\label{sec:traffic_datasets}
Traffic forecasting has been extensively studied using various types of datasets, primarily collected from fixed sensors deployed across transportation networks. These datasets can be broadly categorized into three main types based on their temporal scope and variable composition.

Long-range prediction datasets, such as the ``Traffic" dataset \cite{autoformer_2021}, 
are commonly used as baselines for evaluating forecasting models designed for extended prediction horizons. This dataset serves as a benchmark in studies of models like S-D mamba \cite{s-d-mamba_2025}
and Autoformer \cite{autoformer_2021}, 
which are primarily developed for forecasting applications beyond traffic prediction, including weather patterns and transformer temperatures. Traffic data represents just one application domain within this broader class of long-horizon forecasting problems.

Univariate traffic prediction datasets focus on single measurement types from individual sensors. The PEMS-BAY and METR-LA datasets \cite{li2018dcrnn}
contain only vehicle speed measurements, while PEMS-03 and PEMS-07 datasets \cite{yu2018stgcn} 
provide traffic flow data exclusively. These datasets have been widely used to evaluate models that predict single traffic variables over short to medium time horizons.

Multivariate traffic datasets offer a more comprehensive view of traffic conditions by including multiple measurement types from the same sensors. The PEMS04 and PEMS08 datasets \cite{yu2018stgcn}, 
which we utilize in this study, contain three variables: vehicle speed, traffic flow, and road occupancy. These datasets enable the evaluation of models that can leverage correlations between different traffic measurements for improved prediction accuracy. Additionally, as traffic data is inherently multivariate, these datasets are more representative of real-world traffic conditions than univariate and long-range prediction datasets, making the analysis in this work more comprehensive and relevant to real-world applications.

While all these datasets are collected from fixed sensor networks, there exist alternative data sources such as floating car data (FCD) from GPS-enabled vehicles \cite{t-drive_2013}. 
However, FCD datasets are outside the scope of this work, as our focus is on understanding sparsity patterns in sensor-based multivariate traffic data.

\subsection{Traffic Forecasting Models}

Traffic forecasting models have evolved through several architectural paradigms \cite{11042911}, each offering different trade-offs between computational efficiency and the ability to capture complex spatial-temporal dependencies. 
Recurrent Neural Network (RNN) based approaches, such as the Diffusion Convolutional Recurrent Neural Network (DCRNN) \cite{li2018dcrnn}, 
combine graph convolutions with gated recurrent units to model spatial diffusion and temporal patterns. These models often serve as strong baselines due to their stability in capturing long-term dependencies. Transformer-based architectures like Trafformer \cite{trafformer_2023}
have also been widely adopted, utilizing self-attention mechanisms to capture global dependencies without the sequential constraints of RNNs. Another paradigm involves dynamic Graph Neural Networks (GNNs), 
exemplified by D2STGNN \cite{shao2022d2stgnn}, 
which attempts to learn evolving spatial correlations rather than relying on static adjacency matrices. Finally, State-Space Models (SSMs), powered by the Mamba block \cite{gu2023mamba}, 
have emerged as a high-performance alternative. For instance, the Mamba4Traffic \cite{mamba4traffic} model 
leverages the efficiency of structured state-space representations and is a strong full-data baseline on multivariate traffic benchmarks.

\subsection{Sparsity in Traffic Data}

Data sparsity is a persistent challenge in traffic forecasting \cite{tvt_sparsity_2025, huang2025ojits_missing}, 
and most existing literature addresses this through data imputation. 
Imputation techniques, ranging from simple statistical averages
to complex generative adversarial networks
and graph-based approaches \cite{2025_sparsity_survey}, 
aim to estimate missing values during the pre-processing stage to ensure a complete dataset for model training. While effective for data cleaning, these methods do not address the inherent sensitivity of the forecasting models themselves when faced with missing data at the point of prediction.

Beyond imputation, recent work has examined robustness to missing sensor observations in intelligent transportation systems \cite{10315958}, while others analyze model performance under structured sensor dropout, similarly to this work. Our work distinguishes itself by systematically evaluating how different architectural paradigms respond to increasing levels of inference-time sparsity in multi-channel datasets, providing a benchmark for operational reliability that goes beyond traditional accuracy metrics.

\subsection{Operational Accuracy Thresholds}
\label{sec:accuracy_thresholds}

The utility of traffic performance measures is intrinsically linked to the accuracy requirements of their intended applications. The \textit{Guide to Benchmarking Operations Performance Measures} \cite{2008guide}, prepared for the NCHRP Transportation Research Board of The National Academies, provides a widely recognized framework for characterizing the accuracy levels required to support various transportation applications. The guide establishes acceptable accuracy ranges for performance measures across four primary classes: traffic engineering, transportation planning, traffic management, and traveler information. These ranges are based on input from project participants, pilot test results, and extensive literature reviews, ensuring their practical applicability in the transportation industry.

A critical finding of this framework is that an application's effectiveness is compromised if the error in a performance measure exceeds the specified range. The establishment of these clear benchmarks underscores the necessity for forecasting models to incorporate an awareness of industry-specific accuracy requirements. Consequently, a primary objective in the development of advanced traffic models is to ensure their predictions conform to these established standards, guaranteeing their practical utility. For our study, we have selected a 10\% accuracy threshold as a critical benchmark. This figure represents the common upper bound of acceptable error for vehicle traffic throughput and speed in both traffic management and transportation planning applications. We use this 10\% threshold to evaluate the operational robustness of the selected forecasting models by determining the point at which data sparsity renders model predictions unsuitable for practical deployment.

\section{Methodology}
\label{sec:methodology}

\subsection{Problem Definition}

We formulate sparsity robustness evaluation as measuring prediction degradation under preset random sensor dropout at inference time. The goal is to quantify how traffic forecasting models perform when portions of the sensor network become unavailable, simulating real-world scenarios such as sensor malfunction, communication failures, or maintenance periods.

Given a trained traffic forecasting model $h$ that maps historical observations to future predictions, we evaluate its robustness by systematically removing sensors from the input and measuring the resulting prediction error. This evaluation occurs at inference time without model retraining, isolating each model's sensitivity to missing data.

Traffic data is represented as a 4-dimensional tensor $X \in \mathbb{R}^{m \times t_{in} \times N \times f}$, where $m$ is the batch size, $t_{in}$ is the input time window length, $N$ is the number of sensors/nodes in the traffic network, and $f$ is the feature dimension, containing the traffic variables.

For the PEMS04 and PEMS08 datasets used in our experiments, the three prediction targets are:
\begin{itemize}
\item Traffic Flow (q): The total number of vehicles passing a sensor location in a 5-minute interval.
\item Average Speed (v): The mean speed of vehicles detected during a 5-minute interval, measured in miles per hour.
\item Average Occupancy (o): The fraction of time within a 5-minute interval that the sensor was occupied by vehicles, calculated as o = (Total occupied time / Interval length) $\times 100\%$.
\end{itemize}

The prediction task follows the standard multi-horizon forecasting formulation:
\begin{equation}
[X_{t+1}, X_{t+2}, ..., X_{t+t_{out}}] = h([X_{t-t_{in}+1}, X_{t-t_{in}+2}, ..., X_t])
\label{eq:prediction_function}
\end{equation}

where $h$ denotes the prediction function that takes $m$ samples of traffic measurements from $N$ nodes, with each sample of length $t_{in}$ as input. The model forecasts the future $t_{out}$ steps for all nodes. In our experiments, $t_{in} = t_{out} = 12$, corresponding to 1-hour windows at 5-minute resolution.

\subsection{Sparsity Simulation}
\label{sec:sparsity_simulation}

To simulate sensor unavailability, we define a sparsity level $\rho \in [0,1]$ as the fraction of sensors to be masked. For a network with $N$ sensors, we mask $k = \mathrm{round}(\rho N)$ sensors.

The set of sensor indices is denoted as $S = \{1,2, \ldots, N\}$. For each sparsity level $\rho$, a subset $S_{mask} \subset S$ is sampled uniformly at random such that $|S_{mask}| = k$. The corresponding active set is $S_{active} = S \setminus S_{mask}$, and its cardinality satisfies $|S_{active}| = N - k$. 
We evaluate three independent random masks per sparsity level using fixed seeds 43, 44, and 45. Within each seed, the mask is held fixed for reproducibility.

In operational networks, sensor outages can exhibit spatial correlation, temporal correlation, or other structured failure patterns \cite{rajagopal2009sensorhealth}. We adopt uniform random masking as a general representation of sparsity, as it characterizes the aggregate impact of missing sensor values without imposing a particular outage topology. This characterizes model behavior in the most general case, and more specific correlation structures can be incorporated when the benchmarking framework is applied to a particular deployment, depending on its outage patterns and maintenance and communication infrastructure.

Masked sensors are simulated by overwriting their traffic data input histories with a value corresponding to a raw zero reading. In operational loop networks, sensor outages take several distinct forms, including communication failures, missing samples, constant non-zero outputs, and abnormally high values, as classified in the daily diagnostic states used by the PeMS archive \cite{rajagopal2009sensorhealth}. As part of the preprocessing of the PEMS datasets, some failure modes, such as communication outages or missing sensor data, manifest as erroneous zero values \cite{guo2019astgcn,zuo2023gcnm}, whereas other faults show as high or constant readings. In this analysis, we adopt zero-fill as a general case for sensor outages because it covers a broad class of failure modes, while excluding constant-value and high-value error patterns from the scope of this work. Last-observation carry-forward and mean fill are valid treatments of a silent or failed sensor and remain useful subjects for further study.

To simulate sparsity, given the model input $X \in \mathbb{R}^{m \times t_{in} \times N \times f}$, where $f$ is the number of input features per sensor, masking is applied to the relevant features. In some model pipelines, $f$ includes not only the core traffic variables but also additional covariates appended as extra feature channels, such as time-of-day encodings or day-of-week indicators (e.g., a 7-dimensional one-hot vector). Consequently, $f$ may exceed three even when the forecasting target remains the standard traffic triplet. To make the masking operation consistent across models, we denote by $C$ the number of traffic target channels (here, $C=3$ for flow, speed, and occupancy) and apply masking only to these $C$ channels, leaving any auxiliary covariates unchanged.
For each masked sensor $i \in S_{mask}$, we overwrite its traffic input history across all timesteps with a constant fill value corresponding to a raw zero reading under the same preprocessing used to construct the model inputs. Inputs are z-score normalized per channel using training-set mean $\mu_c$ and standard deviation $\sigma_c$, therefore the fill value is:
\begin{equation}
z_{0,c} = \frac{0 - \mu_c}{\sigma_c} = -\frac{\mu_c}{\sigma_c}
\end{equation}

The masked input tensor $X_{masked}$ is defined as:
\begin{equation}
X_{masked}[:,:,i,c] = \begin{cases}
z_{0,c}, & i \in S_{mask}, c \in \{1, \ldots, C\} \\
X[:,:,i,c], & \text{otherwise} \\
\end{cases}
\end{equation}

This procedure applies a uniform zero fill in the model's normalized input space. For a given sparsity level $\rho$ and seed, the same masked sensor subset $S_{mask}$ is applied across all timesteps and across the full test set to isolate the effect of sensor removal from mask-to-mask variability.

We use two discrete sparsity grids. The complete grid $P_{10}=\{0.0,0.1,\ldots,1.0\}$ is reported in Table~\ref{tab:multilevel_sparsity} and in the ranking matrices. Under the Gaussian weighting introduced below, sparsity levels beyond $35\%$ contribute negligibly to the weighted scores, which motivates the operational grid $P_{5}=\{0.00,0.05,\ldots,0.35\}$ used for breakdown curves, relative-degradation curves, and the weighted aggregates in Table~\ref{tab:sparsity_degradation}. For each $\rho$ on a given grid, the same $S_{mask}(\rho)$ is reused across all evaluated models within each seed. Reported sparsity metrics are averaged over the three masks.

\subsection{Evaluation Metrics}
\label{sec:evaluation_metrics}

To address scale differences across traffic variables, we employ z-score normalization for evaluation. For each channel $c \in \{\text{flow}, \text{speed}, \text{occupancy}\}$, we normalize the raw value $x$ as follows:
\begin{equation}
    \tilde{x} = \frac{x - \mu_c}{\sigma_c}
\end{equation}
where $\tilde{x}$ is the z-score normalized value, and $\mu_c$ and $\sigma_c$ are the mean and standard deviation of channel $c$, computed over the training set. This normalization ensures that error values are comparable across variables with differing units and value ranges.

\subsubsection{Vehicular Error Metrics}
\label{sec:vehicular_error_metrics}

Following the vehicular evaluation framework defined in Mamba4Traffic \cite{mamba4traffic}, 
we report per-channel errors computed in z-score space on the normalized ground truth $\tilde{y}$ and normalized predictions $\tilde{\hat{y}}$. For a given channel $c$ and $N$ evaluation points, the channel-wise vehicular metrics are:
\begin{subequations}
\begin{align}
vMAE_c  &= \frac{1}{N} \sum_{i=1}^{N} \left| \tilde{y}_{i,c} - \tilde{\hat{y}}_{i,c} \right| \\
vRMSE_c &= \sqrt{ \frac{1}{N} \sum_{i=1}^{N} \left( \tilde{y}_{i,c} - \tilde{\hat{y}}_{i,c} \right)^2 }
\end{align}
\end{subequations}

We exclude percentage-based metrics such as MAPE and its normalized variant from this analysis. Under z-score normalization, percentage errors can spike when the normalized ground truth is close to zero, making the values difficult to interpret and less operationally meaningful for comparing robustness under sensor sparsity. We therefore focus on vMAE and vRMSE.

\subsubsection{Channel-Averaged Metrics}

For each sparsity level $\rho$, we compute a single scalar $vMAE_a (\rho)$ by averaging over all test samples, all $t_{out} = 12$ prediction horizons, and all nodes. We then average across the three traffic channels to obtain the channel-averaged score:
\begin{equation}
vMAE_a = \frac{1}{3} (vMAE_{flow} + vMAE_{speed} + vMAE_{occupancy})
\end{equation}
and similarly for $vRMSE_a$.

\subsubsection{Weighted Sweep-Averaged Performance}

While the simple arithmetic mean across the complete grid $P_{10}$ provides a broad overview of model performance, it treats all sparsity levels equally. In practice, real-time traffic management systems must maintain a minimum designed information availability of 90\% for travel-time and speed reporting \cite{cfr511_309}, which corresponds to at most 10\% sensor unavailability under normal operations. An unweighted mean over $P_{10}$ would therefore overweight sparsity regimes that fall outside this operational envelope, including differences in model ranking that emerge only at higher sensor-loss levels. To align our aggregate metrics with expected deployment conditions, we introduce a weighted average over the operational grid $P_{5}$ that emphasizes performance at low sparsity and sharply de-emphasizes contributions beyond 10\% sparsity.

\begin{figure}[!htbp]
    \centering
    \includegraphics[width=0.5\textwidth]{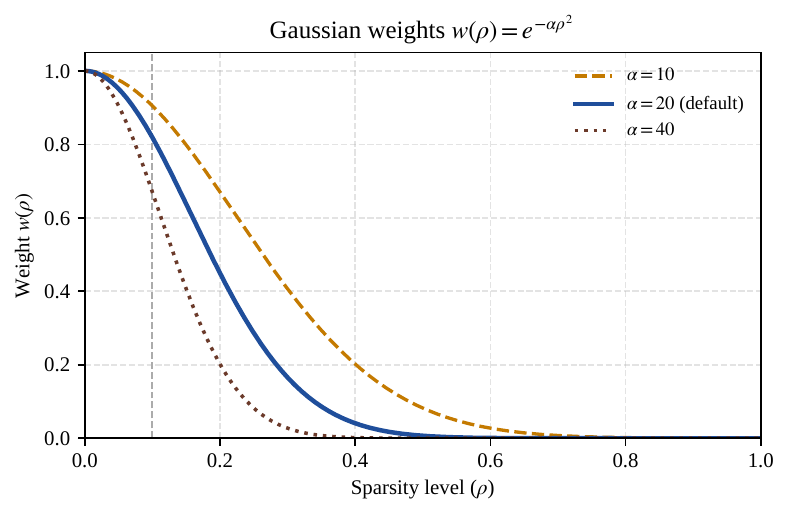}
    \caption{Gaussian weights $w(\rho)=e^{-\alpha\rho^2}$ for $\alpha=10$, $20$ (default), and $40$. The rapid decay beyond $\rho = 0.1$ reflects the 90\% minimum availability requirement for operational real-time traffic systems.}
    \label{fig:weighting_function}
\end{figure}

Specifically, we assign a weight $w(\rho)$ to each sparsity level $\rho \in P_{5}$ based on an assumed probability distribution of sensor failures, illustrated in Figure~\ref{fig:weighting_function}:
\begin{equation}
w(\rho) = e^{-20\rho^2}
\end{equation}

The weights are then normalized to sum to unity:
\begin{equation}
W(\rho) = \frac{w(\rho)}{\sum_{\rho' \in P_{5}} w(\rho')}
\label{eq:weights}
\end{equation}
The weighted sweep-averaged error $\bar{v}MAE_{a}$ is then calculated as:
\begin{equation}
\bar{v}MAE_{a} = \sum_{\rho \in P_{5}} W(\rho) vMAE_a (\rho)
\end{equation}
Under this scheme, $w(0.1)/w(0) = e^{-0.2} \approx 0.82$, whereas $w(0.2)/w(0) = e^{-0.8} \approx 0.45$, so robustness advantages that appear primarily above 10\% sparsity contribute comparatively little to the weighted score. Models are evaluated primarily on stability under data loss within the availability envelope specified for operational ITS systems, rather than on performance under extreme sensor removal that compliant systems are not expected to encounter.

\subsubsection{Relative Degradation Metrics}

To isolate robustness from baseline accuracy, we compute the relative degradation $D(\rho)$ at each sparsity level $\rho$:
\begin{equation}
D(\rho) = \frac{vMAE_a(\rho)}{vMAE_a(0)}
\end{equation}
where $vMAE_a(\rho)$ is the channel-averaged vMAE under $\rho$ sparsity. To express this as a percentage increase relative to the baseline, we define the percentage degradation $D_{\%}(\rho)$:
\begin{equation}
D_{\%}(\rho) = (D(\rho)-1) \times 100
\end{equation}
Under this definition, $D_{\%}(\rho)=0\%$ indicates no change from baseline performance, while $D_{\%}(\rho)=100\%$ indicates that the error has doubled relative to the full-data condition.

To summarize the rate of performance loss on the operational grid, we define the weighted average degradation factor $\bar{D}$ as the weighted mean of the relative degradation over $P_{5}$:
\begin{equation}
\bar{D}_{Metric} = \sum_{\rho \in P_{5}} W(\rho) D(\rho)
\end{equation}
where $W(\rho)$ are the normalized weights defined in Equation \ref{eq:weights}. This metric isolates robustness from baseline accuracy by normalizing errors to the full-data condition. Specifically, while $\bar{v}MAE_a$ reflects absolute performance, $\bar{D}$ characterizes the model's sensitivity to sensor loss, independent of its baseline error.

\subsubsection{Usability Threshold}

We define a breakdown point $\rho^*$ as the smallest sparsity level at which the channel-averaged $vMAE_a$ exceeds a usability threshold $\tau$:
\begin{equation}
\rho^* = \min_{\rho \in P_{5}} \{ \rho : vMAE_a(\rho) > \tau \}
\end{equation}

We derive $\tau$ from a 10\% raw error criterion, as discussed in Section~\ref{sec:accuracy_thresholds} and established in \cite{2008guide}. For channel $c$, the corresponding threshold in $vMAE$ units is:
\begin{equation}
\label{eq:tau}
\tau_{vMAE,c} = \frac{0.10 \mu_c}{\sigma_c}
\end{equation}
To obtain a single usability threshold consistent with the channel-averaged metric, we aggregate across the three traffic channels by taking the mean:
\begin{equation}
\label{eq:usability_threshold}
\tau = \frac{1}{3} (\tau_{vMAE,flow} + \tau_{vMAE,speed} + \tau_{vMAE,occupancy}).
\end{equation}
Here, $\tau$ is the average usability threshold that represents the thresholds of all three traffic variables. Because $vMAE_a (\rho)$ is evaluated on a discrete operational grid $P_{5}$, the breakdown point $\rho^*$ is estimated by linear interpolation between adjacent 5\% samples.

\subsection{Experimental Setup}

All experiments were conducted on a workstation with an AMD EPYC 74F3 24-Core Processor (36 threads), 433 GB of RAM, and an NVIDIA A10-24Q GPU with 24 GB of memory, running Ubuntu 22.04.5 LTS.

\subsection{Experiments}
\label{sec:experiments}

\begin{table*}[!t]
    \caption{Weighted performance metrics and degradation factors for different models on PEMS04 and PEMS08 datasets. Entries show the mean with sample standard deviation over three independent masks at each sparsity level, using $w(\rho)=e^{-20\rho^2}$ over 0--35\% sparsity. Bold indicates best and underline second-best in each column (by mean).}
    \label{tab:sparsity_degradation}
    % Auto-generated by scripts/generate_table1_multiseed.py — do not edit by hand.
% Standardized-input checkpoints; Gaussian weights over 0--35% sparsity.
% Mean with sample standard deviation over mask seeds 43--45.

    \begin{tabular*}{\textwidth}{@{\extracolsep{\fill}}cccccccccc@{}}
    \hline
    \multicolumn{2}{c}{Dataset} & \multicolumn{4}{c}{PEMS04} & \multicolumn{4}{c}{PEMS08} \\ \hline
    Model & Metric & Flow & Occupancy & Speed & \multicolumn{1}{c}{Avg} & Flow & Occupancy & Speed & Avg \\ \hline \\[0.9em]
    \multirow{4}{*}{DCRNN} & $\bar{v}$MAE & \begin{tabular}[t]{@{}c@{}}\underline{0.246} \\[-0.05ex] {\fontsize{6}{7}\selectfont \hspace{0.45em}$\pm$0.001}\end{tabular} & \begin{tabular}[t]{@{}c@{}}0.400 \\[-0.05ex] {\fontsize{6}{7}\selectfont \hspace{0.45em}$\pm$0.001}\end{tabular} & \begin{tabular}[t]{@{}c@{}}0.605 \\[-0.05ex] {\fontsize{6}{7}\selectfont \hspace{0.45em}$\pm$0.004}\end{tabular} & \begin{tabular}[t]{@{}c@{}}0.417 \\[-0.05ex] {\fontsize{6}{7}\selectfont \hspace{0.45em}$\pm$0.001}\end{tabular} & \begin{tabular}[t]{@{}c@{}}0.278 \\[-0.05ex] {\fontsize{6}{7}\selectfont \hspace{0.45em}$\pm$0.003}\end{tabular} & \begin{tabular}[t]{@{}c@{}}0.449 \\[-0.05ex] {\fontsize{6}{7}\selectfont \hspace{0.45em}$\pm$0.005}\end{tabular} & \begin{tabular}[t]{@{}c@{}}0.585 \\[-0.05ex] {\fontsize{6}{7}\selectfont \hspace{0.45em}$\pm$0.009}\end{tabular} & \begin{tabular}[t]{@{}c@{}}0.438 \\[-0.05ex] {\fontsize{6}{7}\selectfont \hspace{0.45em}$\pm$0.003}\end{tabular} \\
    \noalign{\vskip 0.45em}
    & $\bar{D}_{MAE}$ & \begin{tabular}[t]{@{}c@{}}\underline{1.98$\times$} \\[-0.05ex] {\fontsize{6}{7}\selectfont \hspace{0.45em}$\pm$0.006}\end{tabular} & \begin{tabular}[t]{@{}c@{}}\textbf{1.22$\times$} \\[-0.05ex] {\fontsize{6}{7}\selectfont \hspace{0.45em}$\pm$0.003}\end{tabular} & \begin{tabular}[t]{@{}c@{}}3.00$\times$ \\[-0.05ex] {\fontsize{6}{7}\selectfont \hspace{0.45em}$\pm$0.022}\end{tabular} & \begin{tabular}[t]{@{}c@{}}\underline{1.91$\times$} \\[-0.05ex] {\fontsize{6}{7}\selectfont \hspace{0.45em}$\pm$0.005}\end{tabular} & \begin{tabular}[t]{@{}c@{}}2.63$\times$ \\[-0.05ex] {\fontsize{6}{7}\selectfont \hspace{0.45em}$\pm$0.027}\end{tabular} & \begin{tabular}[t]{@{}c@{}}\underline{1.23$\times$} \\[-0.05ex] {\fontsize{6}{7}\selectfont \hspace{0.45em}$\pm$0.014}\end{tabular} & \begin{tabular}[t]{@{}c@{}}2.69$\times$ \\[-0.05ex] {\fontsize{6}{7}\selectfont \hspace{0.45em}$\pm$0.042}\end{tabular} & \begin{tabular}[t]{@{}c@{}}\underline{1.91$\times$} \\[-0.05ex] {\fontsize{6}{7}\selectfont \hspace{0.45em}$\pm$0.014}\end{tabular} \\
    \noalign{\vskip 0.9em}
    & $\bar{v}$RMSE & \begin{tabular}[t]{@{}c@{}}0.426 \\[-0.05ex] {\fontsize{6}{7}\selectfont \hspace{0.45em}$\pm$0.005}\end{tabular} & \begin{tabular}[t]{@{}c@{}}\underline{0.691} \\[-0.05ex] {\fontsize{6}{7}\selectfont \hspace{0.45em}$\pm$0.002}\end{tabular} & \begin{tabular}[t]{@{}c@{}}1.189 \\[-0.05ex] {\fontsize{6}{7}\selectfont \hspace{0.45em}$\pm$0.011}\end{tabular} & \begin{tabular}[t]{@{}c@{}}0.768 \\[-0.05ex] {\fontsize{6}{7}\selectfont \hspace{0.45em}$\pm$0.003}\end{tabular} & \begin{tabular}[t]{@{}c@{}}0.469 \\[-0.05ex] {\fontsize{6}{7}\selectfont \hspace{0.45em}$\pm$0.012}\end{tabular} & \begin{tabular}[t]{@{}c@{}}0.746 \\[-0.05ex] {\fontsize{6}{7}\selectfont \hspace{0.45em}$\pm$0.005}\end{tabular} & \begin{tabular}[t]{@{}c@{}}1.163 \\[-0.05ex] {\fontsize{6}{7}\selectfont \hspace{0.45em}$\pm$0.011}\end{tabular} & \begin{tabular}[t]{@{}c@{}}0.793 \\[-0.05ex] {\fontsize{6}{7}\selectfont \hspace{0.45em}$\pm$0.003}\end{tabular} \\
    \noalign{\vskip 0.45em}
    & $\bar{D}_{RMSE}$ & \begin{tabular}[t]{@{}c@{}}2.12$\times$ \\[-0.05ex] {\fontsize{6}{7}\selectfont \hspace{0.45em}$\pm$0.025}\end{tabular} & \begin{tabular}[t]{@{}c@{}}\textbf{1.12$\times$} \\[-0.05ex] {\fontsize{6}{7}\selectfont \hspace{0.45em}$\pm$0.004}\end{tabular} & \begin{tabular}[t]{@{}c@{}}2.67$\times$ \\[-0.05ex] {\fontsize{6}{7}\selectfont \hspace{0.45em}$\pm$0.024}\end{tabular} & \begin{tabular}[t]{@{}c@{}}\underline{1.83$\times$} \\[-0.05ex] {\fontsize{6}{7}\selectfont \hspace{0.45em}$\pm$0.006}\end{tabular} & \begin{tabular}[t]{@{}c@{}}2.75$\times$ \\[-0.05ex] {\fontsize{6}{7}\selectfont \hspace{0.45em}$\pm$0.072}\end{tabular} & \begin{tabular}[t]{@{}c@{}}\underline{1.16$\times$} \\[-0.05ex] {\fontsize{6}{7}\selectfont \hspace{0.45em}$\pm$0.008}\end{tabular} & \begin{tabular}[t]{@{}c@{}}\underline{2.23$\times$} \\[-0.05ex] {\fontsize{6}{7}\selectfont \hspace{0.45em}$\pm$0.021}\end{tabular} & \begin{tabular}[t]{@{}c@{}}\underline{1.78$\times$} \\[-0.05ex] {\fontsize{6}{7}\selectfont \hspace{0.45em}$\pm$0.007}\end{tabular} \\
    \\[0.1em] \hline \\[0.9em]

    \multirow{4}{*}{D2STGNN} & $\bar{v}$MAE & \begin{tabular}[t]{@{}c@{}}0.476 \\[-0.05ex] {\fontsize{6}{7}\selectfont \hspace{0.45em}$\pm$0.022}\end{tabular} & \begin{tabular}[t]{@{}c@{}}1.023 \\[-0.05ex] {\fontsize{6}{7}\selectfont \hspace{0.45em}$\pm$0.020}\end{tabular} & \begin{tabular}[t]{@{}c@{}}0.813 \\[-0.05ex] {\fontsize{6}{7}\selectfont \hspace{0.45em}$\pm$0.001}\end{tabular} & \begin{tabular}[t]{@{}c@{}}0.771 \\[-0.05ex] {\fontsize{6}{7}\selectfont \hspace{0.45em}$\pm$0.014}\end{tabular} & \begin{tabular}[t]{@{}c@{}}\underline{0.197} \\[-0.05ex] {\fontsize{6}{7}\selectfont \hspace{0.45em}$\pm$0.004}\end{tabular} & \begin{tabular}[t]{@{}c@{}}\textbf{0.279} \\[-0.05ex] {\fontsize{6}{7}\selectfont \hspace{0.45em}$\pm$0.001}\end{tabular} & \begin{tabular}[t]{@{}c@{}}0.659 \\[-0.05ex] {\fontsize{6}{7}\selectfont \hspace{0.45em}$\pm$0.002}\end{tabular} & \begin{tabular}[t]{@{}c@{}}\underline{0.378} \\[-0.05ex] {\fontsize{6}{7}\selectfont \hspace{0.45em}$\pm$0.001}\end{tabular} \\
    \noalign{\vskip 0.45em}
    & $\bar{D}_{MAE}$ & \begin{tabular}[t]{@{}c@{}}4.02$\times$ \\[-0.05ex] {\fontsize{6}{7}\selectfont \hspace{0.45em}$\pm$0.187}\end{tabular} & \begin{tabular}[t]{@{}c@{}}6.07$\times$ \\[-0.05ex] {\fontsize{6}{7}\selectfont \hspace{0.45em}$\pm$0.121}\end{tabular} & \begin{tabular}[t]{@{}c@{}}4.26$\times$ \\[-0.05ex] {\fontsize{6}{7}\selectfont \hspace{0.45em}$\pm$0.005}\end{tabular} & \begin{tabular}[t]{@{}c@{}}4.84$\times$ \\[-0.05ex] {\fontsize{6}{7}\selectfont \hspace{0.45em}$\pm$0.089}\end{tabular} & \begin{tabular}[t]{@{}c@{}}2.07$\times$ \\[-0.05ex] {\fontsize{6}{7}\selectfont \hspace{0.45em}$\pm$0.041}\end{tabular} & \begin{tabular}[t]{@{}c@{}}1.58$\times$ \\[-0.05ex] {\fontsize{6}{7}\selectfont \hspace{0.45em}$\pm$0.005}\end{tabular} & \begin{tabular}[t]{@{}c@{}}3.28$\times$ \\[-0.05ex] {\fontsize{6}{7}\selectfont \hspace{0.45em}$\pm$0.010}\end{tabular} & \begin{tabular}[t]{@{}c@{}}2.40$\times$ \\[-0.05ex] {\fontsize{6}{7}\selectfont \hspace{0.45em}$\pm$0.009}\end{tabular} \\
    \noalign{\vskip 0.9em}
    & $\bar{v}$RMSE & \begin{tabular}[t]{@{}c@{}}0.639 \\[-0.05ex] {\fontsize{6}{7}\selectfont \hspace{0.45em}$\pm$0.028}\end{tabular} & \begin{tabular}[t]{@{}c@{}}1.578 \\[-0.05ex] {\fontsize{6}{7}\selectfont \hspace{0.45em}$\pm$0.013}\end{tabular} & \begin{tabular}[t]{@{}c@{}}1.477 \\[-0.05ex] {\fontsize{6}{7}\selectfont \hspace{0.45em}$\pm$0.013}\end{tabular} & \begin{tabular}[t]{@{}c@{}}1.231 \\[-0.05ex] {\fontsize{6}{7}\selectfont \hspace{0.45em}$\pm$0.008}\end{tabular} & \begin{tabular}[t]{@{}c@{}}\underline{0.390} \\[-0.05ex] {\fontsize{6}{7}\selectfont \hspace{0.45em}$\pm$0.012}\end{tabular} & \begin{tabular}[t]{@{}c@{}}\textbf{0.589} \\[-0.05ex] {\fontsize{6}{7}\selectfont \hspace{0.45em}$\pm$0.005}\end{tabular} & \begin{tabular}[t]{@{}c@{}}1.472 \\[-0.05ex] {\fontsize{6}{7}\selectfont \hspace{0.45em}$\pm$0.011}\end{tabular} & \begin{tabular}[t]{@{}c@{}}0.817 \\[-0.05ex] {\fontsize{6}{7}\selectfont \hspace{0.45em}$\pm$0.008}\end{tabular} \\
    \noalign{\vskip 0.45em}
    & $\bar{D}_{RMSE}$ & \begin{tabular}[t]{@{}c@{}}3.17$\times$ \\[-0.05ex] {\fontsize{6}{7}\selectfont \hspace{0.45em}$\pm$0.141}\end{tabular} & \begin{tabular}[t]{@{}c@{}}3.79$\times$ \\[-0.05ex] {\fontsize{6}{7}\selectfont \hspace{0.45em}$\pm$0.031}\end{tabular} & \begin{tabular}[t]{@{}c@{}}3.43$\times$ \\[-0.05ex] {\fontsize{6}{7}\selectfont \hspace{0.45em}$\pm$0.029}\end{tabular} & \begin{tabular}[t]{@{}c@{}}3.53$\times$ \\[-0.05ex] {\fontsize{6}{7}\selectfont \hspace{0.45em}$\pm$0.023}\end{tabular} & \begin{tabular}[t]{@{}c@{}}2.47$\times$ \\[-0.05ex] {\fontsize{6}{7}\selectfont \hspace{0.45em}$\pm$0.076}\end{tabular} & \begin{tabular}[t]{@{}c@{}}1.36$\times$ \\[-0.05ex] {\fontsize{6}{7}\selectfont \hspace{0.45em}$\pm$0.012}\end{tabular} & \begin{tabular}[t]{@{}c@{}}3.01$\times$ \\[-0.05ex] {\fontsize{6}{7}\selectfont \hspace{0.45em}$\pm$0.023}\end{tabular} & \begin{tabular}[t]{@{}c@{}}2.27$\times$ \\[-0.05ex] {\fontsize{6}{7}\selectfont \hspace{0.45em}$\pm$0.021}\end{tabular} \\
    \\[0.1em] \hline \\[0.9em]

    \multirow{4}{*}{Mamba4Traffic} & $\bar{v}$MAE & \begin{tabular}[t]{@{}c@{}}\textbf{0.193} \\[-0.05ex] {\fontsize{6}{7}\selectfont \hspace{0.45em}$\pm$0.003}\end{tabular} & \begin{tabular}[t]{@{}c@{}}\textbf{0.244} \\[-0.05ex] {\fontsize{6}{7}\selectfont \hspace{0.45em}$\pm$0.001}\end{tabular} & \begin{tabular}[t]{@{}c@{}}\underline{0.548} \\[-0.05ex] {\fontsize{6}{7}\selectfont \hspace{0.45em}$\pm$0.004}\end{tabular} & \begin{tabular}[t]{@{}c@{}}\textbf{0.328} \\[-0.05ex] {\fontsize{6}{7}\selectfont \hspace{0.45em}$\pm$0.001}\end{tabular} & \begin{tabular}[t]{@{}c@{}}\textbf{0.177} \\[-0.05ex] {\fontsize{6}{7}\selectfont \hspace{0.45em}$\pm$0.003}\end{tabular} & \begin{tabular}[t]{@{}c@{}}\underline{0.289} \\[-0.05ex] {\fontsize{6}{7}\selectfont \hspace{0.45em}$\pm$0.003}\end{tabular} & \begin{tabular}[t]{@{}c@{}}\underline{0.529} \\[-0.05ex] {\fontsize{6}{7}\selectfont \hspace{0.45em}$\pm$0.007}\end{tabular} & \begin{tabular}[t]{@{}c@{}}\textbf{0.332} \\[-0.05ex] {\fontsize{6}{7}\selectfont \hspace{0.45em}$\pm$0.002}\end{tabular} \\
    \noalign{\vskip 0.45em}
    & $\bar{D}_{MAE}$ & \begin{tabular}[t]{@{}c@{}}\textbf{1.57$\times$} \\[-0.05ex] {\fontsize{6}{7}\selectfont \hspace{0.45em}$\pm$0.023}\end{tabular} & \begin{tabular}[t]{@{}c@{}}\underline{1.31$\times$} \\[-0.05ex] {\fontsize{6}{7}\selectfont \hspace{0.45em}$\pm$0.003}\end{tabular} & \begin{tabular}[t]{@{}c@{}}\underline{2.86$\times$} \\[-0.05ex] {\fontsize{6}{7}\selectfont \hspace{0.45em}$\pm$0.021}\end{tabular} & \begin{tabular}[t]{@{}c@{}}1.97$\times$ \\[-0.05ex] {\fontsize{6}{7}\selectfont \hspace{0.45em}$\pm$0.005}\end{tabular} & \begin{tabular}[t]{@{}c@{}}\textbf{1.81$\times$} \\[-0.05ex] {\fontsize{6}{7}\selectfont \hspace{0.45em}$\pm$0.028}\end{tabular} & \begin{tabular}[t]{@{}c@{}}1.56$\times$ \\[-0.05ex] {\fontsize{6}{7}\selectfont \hspace{0.45em}$\pm$0.016}\end{tabular} & \begin{tabular}[t]{@{}c@{}}\underline{2.56$\times$} \\[-0.05ex] {\fontsize{6}{7}\selectfont \hspace{0.45em}$\pm$0.035}\end{tabular} & \begin{tabular}[t]{@{}c@{}}2.03$\times$ \\[-0.05ex] {\fontsize{6}{7}\selectfont \hspace{0.45em}$\pm$0.013}\end{tabular} \\
    \noalign{\vskip 0.9em}
    & $\bar{v}$RMSE & \begin{tabular}[t]{@{}c@{}}\textbf{0.359} \\[-0.05ex] {\fontsize{6}{7}\selectfont \hspace{0.45em}$\pm$0.008}\end{tabular} & \begin{tabular}[t]{@{}c@{}}\textbf{0.530} \\[-0.05ex] {\fontsize{6}{7}\selectfont \hspace{0.45em}$\pm$0.002}\end{tabular} & \begin{tabular}[t]{@{}c@{}}\underline{1.157} \\[-0.05ex] {\fontsize{6}{7}\selectfont \hspace{0.45em}$\pm$0.014}\end{tabular} & \begin{tabular}[t]{@{}c@{}}\underline{0.682} \\[-0.05ex] {\fontsize{6}{7}\selectfont \hspace{0.45em}$\pm$0.003}\end{tabular} & \begin{tabular}[t]{@{}c@{}}\textbf{0.351} \\[-0.05ex] {\fontsize{6}{7}\selectfont \hspace{0.45em}$\pm$0.008}\end{tabular} & \begin{tabular}[t]{@{}c@{}}\underline{0.644} \\[-0.05ex] {\fontsize{6}{7}\selectfont \hspace{0.45em}$\pm$0.011}\end{tabular} & \begin{tabular}[t]{@{}c@{}}\underline{1.143} \\[-0.05ex] {\fontsize{6}{7}\selectfont \hspace{0.45em}$\pm$0.022}\end{tabular} & \begin{tabular}[t]{@{}c@{}}\underline{0.712} \\[-0.05ex] {\fontsize{6}{7}\selectfont \hspace{0.45em}$\pm$0.008}\end{tabular} \\
    \noalign{\vskip 0.45em}
    & $\bar{D}_{RMSE}$ & \begin{tabular}[t]{@{}c@{}}\textbf{1.70$\times$} \\[-0.05ex] {\fontsize{6}{7}\selectfont \hspace{0.45em}$\pm$0.036}\end{tabular} & \begin{tabular}[t]{@{}c@{}}\underline{1.18$\times$} \\[-0.05ex] {\fontsize{6}{7}\selectfont \hspace{0.45em}$\pm$0.005}\end{tabular} & \begin{tabular}[t]{@{}c@{}}\underline{2.64$\times$} \\[-0.05ex] {\fontsize{6}{7}\selectfont \hspace{0.45em}$\pm$0.031}\end{tabular} & \begin{tabular}[t]{@{}c@{}}1.86$\times$ \\[-0.05ex] {\fontsize{6}{7}\selectfont \hspace{0.45em}$\pm$0.008}\end{tabular} & \begin{tabular}[t]{@{}c@{}}\underline{2.12$\times$} \\[-0.05ex] {\fontsize{6}{7}\selectfont \hspace{0.45em}$\pm$0.047}\end{tabular} & \begin{tabular}[t]{@{}c@{}}1.43$\times$ \\[-0.05ex] {\fontsize{6}{7}\selectfont \hspace{0.45em}$\pm$0.026}\end{tabular} & \begin{tabular}[t]{@{}c@{}}2.23$\times$ \\[-0.05ex] {\fontsize{6}{7}\selectfont \hspace{0.45em}$\pm$0.044}\end{tabular} & \begin{tabular}[t]{@{}c@{}}1.90$\times$ \\[-0.05ex] {\fontsize{6}{7}\selectfont \hspace{0.45em}$\pm$0.020}\end{tabular} \\
    \\[0.1em] \hline \\[0.9em]

    \multirow{4}{*}{Trafformer} & $\bar{v}$MAE & \begin{tabular}[t]{@{}c@{}}0.298 \\[-0.05ex] {\fontsize{6}{7}\selectfont \hspace{0.45em}$\pm$0.005}\end{tabular} & \begin{tabular}[t]{@{}c@{}}\underline{0.400} \\[-0.05ex] {\fontsize{6}{7}\selectfont \hspace{0.45em}$\pm$0.004}\end{tabular} & \begin{tabular}[t]{@{}c@{}}\textbf{0.523} \\[-0.05ex] {\fontsize{6}{7}\selectfont \hspace{0.45em}$\pm$0.004}\end{tabular} & \begin{tabular}[t]{@{}c@{}}\underline{0.407} \\[-0.05ex] {\fontsize{6}{7}\selectfont \hspace{0.45em}$\pm$0.004}\end{tabular} & \begin{tabular}[t]{@{}c@{}}0.266 \\[-0.05ex] {\fontsize{6}{7}\selectfont \hspace{0.45em}$\pm$0.002}\end{tabular} & \begin{tabular}[t]{@{}c@{}}0.400 \\[-0.05ex] {\fontsize{6}{7}\selectfont \hspace{0.45em}$\pm$0.001}\end{tabular} & \begin{tabular}[t]{@{}c@{}}\textbf{0.486} \\[-0.05ex] {\fontsize{6}{7}\selectfont \hspace{0.45em}$\pm$0.007}\end{tabular} & \begin{tabular}[t]{@{}c@{}}0.384 \\[-0.05ex] {\fontsize{6}{7}\selectfont \hspace{0.45em}$\pm$0.003}\end{tabular} \\
    \noalign{\vskip 0.45em}
    & $\bar{D}_{MAE}$ & \begin{tabular}[t]{@{}c@{}}2.08$\times$ \\[-0.05ex] {\fontsize{6}{7}\selectfont \hspace{0.45em}$\pm$0.036}\end{tabular} & \begin{tabular}[t]{@{}c@{}}1.47$\times$ \\[-0.05ex] {\fontsize{6}{7}\selectfont \hspace{0.45em}$\pm$0.014}\end{tabular} & \begin{tabular}[t]{@{}c@{}}\textbf{1.93$\times$} \\[-0.05ex] {\fontsize{6}{7}\selectfont \hspace{0.45em}$\pm$0.014}\end{tabular} & \begin{tabular}[t]{@{}c@{}}\textbf{1.78$\times$} \\[-0.05ex] {\fontsize{6}{7}\selectfont \hspace{0.45em}$\pm$0.019}\end{tabular} & \begin{tabular}[t]{@{}c@{}}\underline{2.05$\times$} \\[-0.05ex] {\fontsize{6}{7}\selectfont \hspace{0.45em}$\pm$0.018}\end{tabular} & \begin{tabular}[t]{@{}c@{}}\textbf{1.19$\times$} \\[-0.05ex] {\fontsize{6}{7}\selectfont \hspace{0.45em}$\pm$0.003}\end{tabular} & \begin{tabular}[t]{@{}c@{}}\textbf{1.56$\times$} \\[-0.05ex] {\fontsize{6}{7}\selectfont \hspace{0.45em}$\pm$0.023}\end{tabular} & \begin{tabular}[t]{@{}c@{}}\textbf{1.48$\times$} \\[-0.05ex] {\fontsize{6}{7}\selectfont \hspace{0.45em}$\pm$0.011}\end{tabular} \\
    \noalign{\vskip 0.9em}
    & $\bar{v}$RMSE & \begin{tabular}[t]{@{}c@{}}\underline{0.418} \\[-0.05ex] {\fontsize{6}{7}\selectfont \hspace{0.45em}$\pm$0.006}\end{tabular} & \begin{tabular}[t]{@{}c@{}}0.713 \\[-0.05ex] {\fontsize{6}{7}\selectfont \hspace{0.45em}$\pm$0.004}\end{tabular} & \begin{tabular}[t]{@{}c@{}}\textbf{0.904} \\[-0.05ex] {\fontsize{6}{7}\selectfont \hspace{0.45em}$\pm$0.006}\end{tabular} & \begin{tabular}[t]{@{}c@{}}\textbf{0.678} \\[-0.05ex] {\fontsize{6}{7}\selectfont \hspace{0.45em}$\pm$0.006}\end{tabular} & \begin{tabular}[t]{@{}c@{}}0.406 \\[-0.05ex] {\fontsize{6}{7}\selectfont \hspace{0.45em}$\pm$0.006}\end{tabular} & \begin{tabular}[t]{@{}c@{}}0.741 \\[-0.05ex] {\fontsize{6}{7}\selectfont \hspace{0.45em}$\pm$0.002}\end{tabular} & \begin{tabular}[t]{@{}c@{}}\textbf{0.905} \\[-0.05ex] {\fontsize{6}{7}\selectfont \hspace{0.45em}$\pm$0.008}\end{tabular} & \begin{tabular}[t]{@{}c@{}}\textbf{0.684} \\[-0.05ex] {\fontsize{6}{7}\selectfont \hspace{0.45em}$\pm$0.004}\end{tabular} \\
    \noalign{\vskip 0.45em}
    & $\bar{D}_{RMSE}$ & \begin{tabular}[t]{@{}c@{}}\underline{1.73$\times$} \\[-0.05ex] {\fontsize{6}{7}\selectfont \hspace{0.45em}$\pm$0.027}\end{tabular} & \begin{tabular}[t]{@{}c@{}}1.22$\times$ \\[-0.05ex] {\fontsize{6}{7}\selectfont \hspace{0.45em}$\pm$0.008}\end{tabular} & \begin{tabular}[t]{@{}c@{}}\textbf{1.52$\times$} \\[-0.05ex] {\fontsize{6}{7}\selectfont \hspace{0.45em}$\pm$0.010}\end{tabular} & \begin{tabular}[t]{@{}c@{}}\textbf{1.43$\times$} \\[-0.05ex] {\fontsize{6}{7}\selectfont \hspace{0.45em}$\pm$0.012}\end{tabular} & \begin{tabular}[t]{@{}c@{}}\textbf{1.89$\times$} \\[-0.05ex] {\fontsize{6}{7}\selectfont \hspace{0.45em}$\pm$0.026}\end{tabular} & \begin{tabular}[t]{@{}c@{}}\textbf{1.11$\times$} \\[-0.05ex] {\fontsize{6}{7}\selectfont \hspace{0.45em}$\pm$0.003}\end{tabular} & \begin{tabular}[t]{@{}c@{}}\textbf{1.24$\times$} \\[-0.05ex] {\fontsize{6}{7}\selectfont \hspace{0.45em}$\pm$0.011}\end{tabular} & \begin{tabular}[t]{@{}c@{}}\textbf{1.27$\times$} \\[-0.05ex] {\fontsize{6}{7}\selectfont \hspace{0.45em}$\pm$0.008}\end{tabular} \\
    \\[0.1em] \hline

    \end{tabular*}

    \end{table*}

We conduct two main experiments:
\begin{enumerate}
    \item \textbf{Sparsity robustness evaluation:} All models were trained to convergence on complete data following their multichannel prediction setups defined in \cite{mamba4traffic}. The same standardized scaler z-score normalizes each traffic channel with the training-split $\mu_c$ and $\sigma_c$ before the data are input to the model. Robustness to input sparsity was assessed strictly at inference time, without any retraining or architectural modification, by applying the masking procedure to the model inputs and running the full test set through the same pretrained checkpoint for each sparsity level $\rho$. We report:
    \begin{itemize}
        \item The weighted sweep-averaged errors $\bar{v}MAE_a$ and $\bar{v}RMSE_a$ over the operational grid $P_{5}$, for each channel and overall
        \item The degradation factors $\bar{D}_{MAE_c}$ and $\bar{D}_{RMSE_c}$ and the channel-averaged $\bar{D}_{MAE}$ and $\bar{D}_{RMSE}$, relative to the full-data baseline
        \item The channel-averaged errors $vMAE_a(\rho)$ and $vRMSE_a(\rho)$ at each level of the complete grid $P_{10}$
    \end{itemize}
    Absolute $vMAE_a(\rho)$ is reported on the complete grid $P_{10}$ (11 discrete levels from $0\%$ to $100\%$). Weighted aggregates and breakdown curves use the operational grid $P_{5}$ ($0$--$35\%$ at $5\%$ steps).
    \item \textbf{Per-horizon analysis:} For each prediction horizon $h \in \{1, 2, \ldots, 12\}$, we compute $vMAE_a$ separately at each horizon and sparsity level, enabling a detailed examination of horizon-dependent sensitivity to input sparsity.
\end{enumerate}

\subsection{Dataset Statistics}

We evaluate our models on the PEMS04 and PEMS08 datasets \cite{yu2018stgcn}, 
which provide multi-channel traffic data (speed, flow, occupancy) for comprehensive evaluation. These widely-recognized benchmarks enable direct comparison with existing methods. The dataset sizes and general statistics are shown in Table \ref{tab:dataset_overview}. The statistics that are used for z-score normalization during training and for computing the normalized zero value $z_{0,c}$ during sparsity evaluation are outlined in Table \ref{tab:dataset_stats_detailed}.

\begin{table}[!htbp]
\caption{Dataset Overview}
\label{tab:dataset_overview}
\setlength{\tabcolsep}{3.5pt}
\begin{tabular}{lccccc}
\toprule
Dataset & Sensors & Time Steps & Time Range & Interval & Features \\
\midrule
PEMS04 & 307 & 16,992 & 1/18--2/18 & 5 min & \makecell{Speed,\\Flow,\\Occupancy} \\
PEMS08 & 170 & 17,856 & 7/16--8/16 & 5 min & \makecell{Speed,\\Flow,\\Occupancy} \\
\bottomrule
\end{tabular}
\end{table}

\begin{table}[!htbp]
\centering
\caption{Detailed channel statistics for the PEMS04 and PEMS08 datasets.}
\label{tab:dataset_stats_detailed}
\begin{tabular}{llcccc}
\toprule
Dataset & Channel & Mean ($\mu$) & Std ($\sigma$) & Min & Max \\
\midrule
\multirow{3}{*}{PEMS04} & Flow & 208.85 & 157.33 & 0.00 & 919.00 \\
                        & Occupancy & 0.0520 & 0.0488 & 0.00 & 0.7716 \\
                        & Speed & 63.44 & 8.25 & 3.00 & 85.20 \\
\midrule
\multirow{3}{*}{PEMS08} & Flow & 230.24 & 145.89 & 0.00 & 1147.00 \\
                        & Occupancy & 0.0646 & 0.0450 & 0.00 & 0.8955 \\
                        & Speed & 63.78 & 6.56 & 3.00 & 82.30 \\
\bottomrule
\end{tabular}
\end{table}

\subsection{Baseline Models}

We compare a comprehensive set of baseline models representing diverse architectural approaches. The baseline models span multiple architectural paradigms, including graph neural networks, transformers, recurrent networks, and state space models. The model names and approaches are outlined in Table \ref{tab:baseline_models}.

\begin{table}[!htbp]
\centering
\caption{Baseline Models}
\label{tab:baseline_models}
\begin{tabular}{ll}
\toprule
Model & Modeling Approach \\
\midrule
DCRNN         & GNN + RNN \\
D2STGNN       & Dynamic GNN \\
Trafformer    & Transformer \\
Mamba4Traffic & State-Space Model \\
\bottomrule
\end{tabular}
\end{table}

\section{Results and Analysis}
\label{sec:results_and_analysis}

  \begin{table*}[!t]
  \caption{Channel-averaged vehicular MAE ($vMAE_a$) across sparsity levels for each model on PEMS04 and PEMS08. Values are means over three mask seeds (43, 44, and 45). Best and second-best per column are bold and underlined.}
  \label{tab:multilevel_sparsity}
  % Auto-generated by scripts/generate_multiseed_tables_figures.py — do not edit by hand.
% Channel-averaged vMAE_a mean over mask seeds 43--45; paper grid 0--100% at 10% steps.

  \begin{tabular*}{\textwidth}{@{\extracolsep{\fill}}ccccccccccccc@{}}
  \hline
  Dataset   & Model          & 0\%    & 10\%   & 20\%   & 30\%   & 40\%   & 50\%   & 60\%   & 70\%   & 80\%   & 90\%   & 100\% \\
  \hline
  \noalign{\vskip 0.4em}
  \multirow{4}{*}{PEMS04} & DCRNN          & 0.218 & \underline{0.408} & 0.589 & 0.766 & 0.933 & 1.107 & 1.242 & 1.392 & 1.545 & 1.684 & 1.816 \\
  & D2STGNN        & \textbf{0.159} & 0.666 & 1.347 & 2.177 & 2.685 & 3.078 & 3.342 & 3.609 & 3.794 & 3.956 & 4.097 \\
  & Mamba4Traffic  & \underline{0.167} & \textbf{0.320} & \textbf{0.466} & \textbf{0.637} & \textbf{0.788} & \textbf{0.949} & \underline{1.096} & \underline{1.264} & \underline{1.414} & \underline{1.561} & \underline{1.721} \\
  & Trafformer     & 0.229 & 0.412 & \underline{0.529} & \underline{0.680} & \underline{0.857} & \underline{0.951} & \textbf{1.060} & \textbf{1.161} & \textbf{1.218} & \textbf{1.283} & \textbf{1.335} \\
  \noalign{\vskip 0.4em}
  \hline
  \noalign{\vskip 0.4em}
  \multirow{4}{*}{PEMS08} & DCRNN          & 0.229 & 0.433 & 0.644 & 0.785 & 0.934 & 1.051 & 1.133 & 1.201 & 1.289 & 1.361 & 1.433 \\
  & D2STGNN        & \textbf{0.158} & \underline{0.391} & 0.572 & 0.706 & 0.805 & \underline{0.921} & \underline{0.980} & \underline{1.117} & \underline{1.221} & \underline{1.305} & \underline{1.365} \\
  & Mamba4Traffic  & \underline{0.164} & \textbf{0.316} & \underline{0.487} & \underline{0.638} & \underline{0.797} & 0.960 & 1.097 & 1.269 & 1.424 & 1.585 & 1.766 \\
  & Trafformer     & 0.259 & 0.394 & \textbf{0.485} & \textbf{0.533} & \textbf{0.576} & \textbf{0.642} & \textbf{0.659} & \textbf{0.736} & \textbf{0.781} & \textbf{0.835} & \textbf{0.928} \\
  \noalign{\vskip 0.4em}
  \hline
  \end{tabular*}

\end{table*}

\subsection{Sparsity robustness evaluation}

\begin{figure*}[!t]
    \centering
    \includegraphics[width=\textwidth]{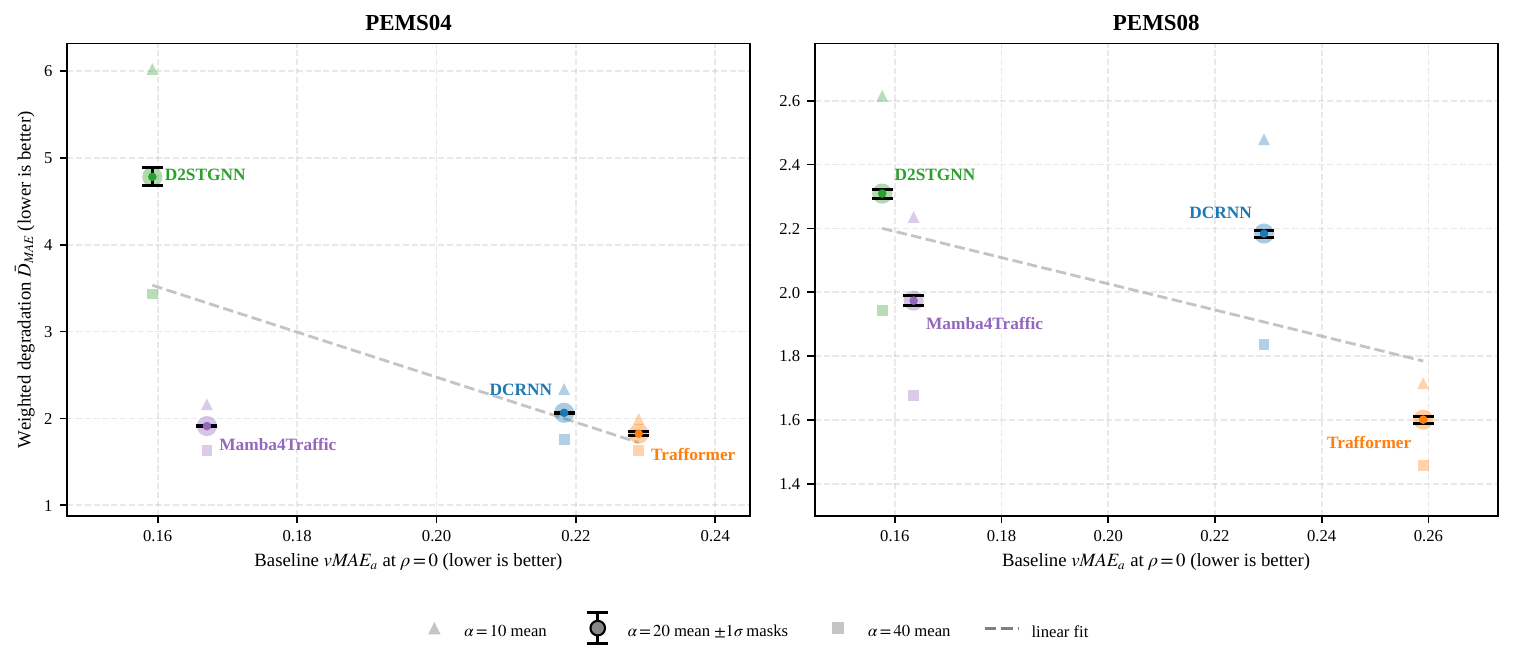}
    \caption{Baseline $vMAE_a$ at $\rho=0$ versus weighted degradation $\bar{D}_{MAE}$ for (a) PEMS04 and (b) PEMS08. Lower values are preferred on both axes.}
    \label{fig:accuracy_robustness_tradeoff}
\end{figure*}

Table~\ref{tab:sparsity_degradation} summarizes the weighted average vehicular metrics and degradation rates across the 0--35\% operational grid.
When models are ranked by the weighted mean degradation factor $\overline{D}$ for vMAE over the 0--35\% operational grid, as seen in Figure~\ref{fig:accuracy_robustness_tradeoff}, Trafformer is the most robust model on both datasets ($1.78\times$ on PEMS04 and $1.48\times$ on PEMS08). DCRNN is second on both datasets ($1.91\times$). Mamba4Traffic is close behind DCRNN ($1.97\times$ and $2.03\times$). D2STGNN, which has the best full-data accuracy, is the least stable model on this metric ($4.84\times$ on PEMS04 and $2.40\times$ on PEMS08). Cross-seed standard deviations on channel-averaged $\bar{D}_{MAE}$ remain small, with D2STGNN on PEMS04 the largest ($\pm 0.089\times$).

Table~\ref{tab:sparsity_degradation} reports vMAE and vRMSE only. Trafformer keeps the lowest channel-averaged $\bar{D}_{MAE}$ on both datasets ($1.78\times$ on PEMS04 and $1.48\times$ on PEMS08). On PEMS04 it also has the lowest speed-channel degradation ($1.93\times$). On PEMS08 it has the lowest occupancy and speed degradation. D2STGNN's degradation is severe on all three PEMS04 channels (occupancy $6.07\times$). DCRNN's occupancy degradation remains comparatively small ($1.22\times$ / $1.23\times$), while its speed channel is the weak point on PEMS04 ($3.00\times$). D2STGNN's PEMS08 sensitivity is likewise largest on speed ($3.28\times$), not flow. D2STGNN still degrades less on PEMS08 than on PEMS04 ($2.40\times$ versus $4.84\times$). No channel-wise $\bar{D}_{MAE}$ falls below $1.0$.

Overall, Table~\ref{tab:sparsity_degradation} reinforces the trade-off between baseline accuracy and robustness under sensor sparsity. Strong baseline performance does not necessarily guarantee robustness when sensors are masked, emphasizing the need to evaluate models within the context of anticipated sensing regimes.

Table~\ref{tab:multilevel_sparsity} enables a direct comparison of operational performance, measured in $vMAE_a$, at selected sparsity levels. At $\rho = 0\%$, D2STGNN achieves the lowest $vMAE_a$ on both datasets ($0.159$ on PEMS04 and $0.158$ on PEMS08), with Mamba4Traffic a close second ($0.167$ and $0.164$). That full-data ordering does not survive modest masking. On PEMS04, Mamba4Traffic leads from $\rho = 10\%$ through $\rho = 50\%$ (including $0.320$ at $10\%$), and Trafformer leads from $\rho = 60\%$ through $\rho = 100\%$. On PEMS08, Mamba4Traffic leads at $\rho = 10\%$ ($0.316$), and Trafformer leads from $\rho = 20\%$ through $\rho = 100\%$. D2STGNN falls to last place from $\rho = 10\%$ onward on PEMS04; on PEMS08 it remains second at $10\%$ and again from $50\%$ onward, but it is never first once sensors are removed.

When 10--20\% of detectors may be unavailable, robustness should weigh as heavily as peak accuracy. A model chosen only by full-data error can become suboptimal after modest masking.

\begin{figure*}[!t]
    \centering
    \subfloat[PEMS04]{\label{fig:ranking_matrix_PEMS04}
        \includegraphics[width=0.47\textwidth]{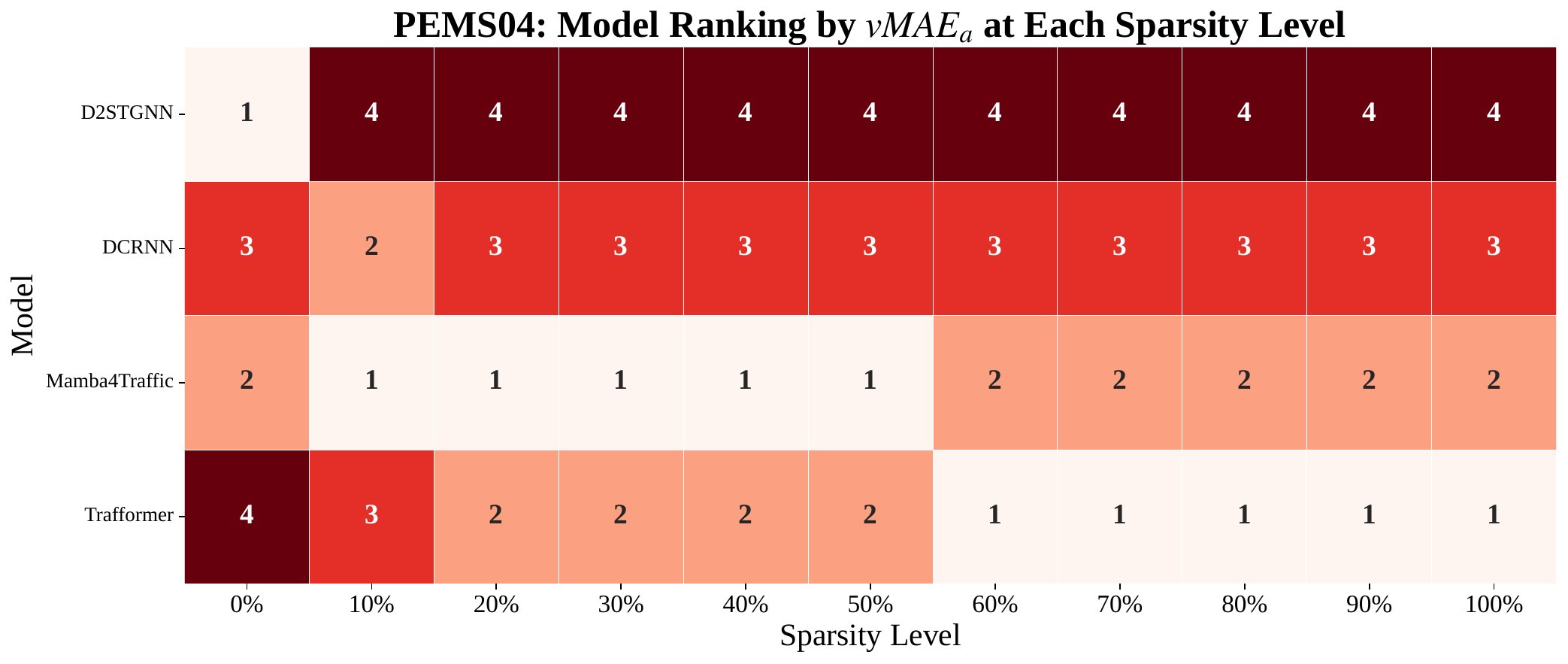}
    }
    \hfill
    \subfloat[PEMS08]{\label{fig:ranking_matrix_PEMS08}
        \includegraphics[width=0.47\textwidth]{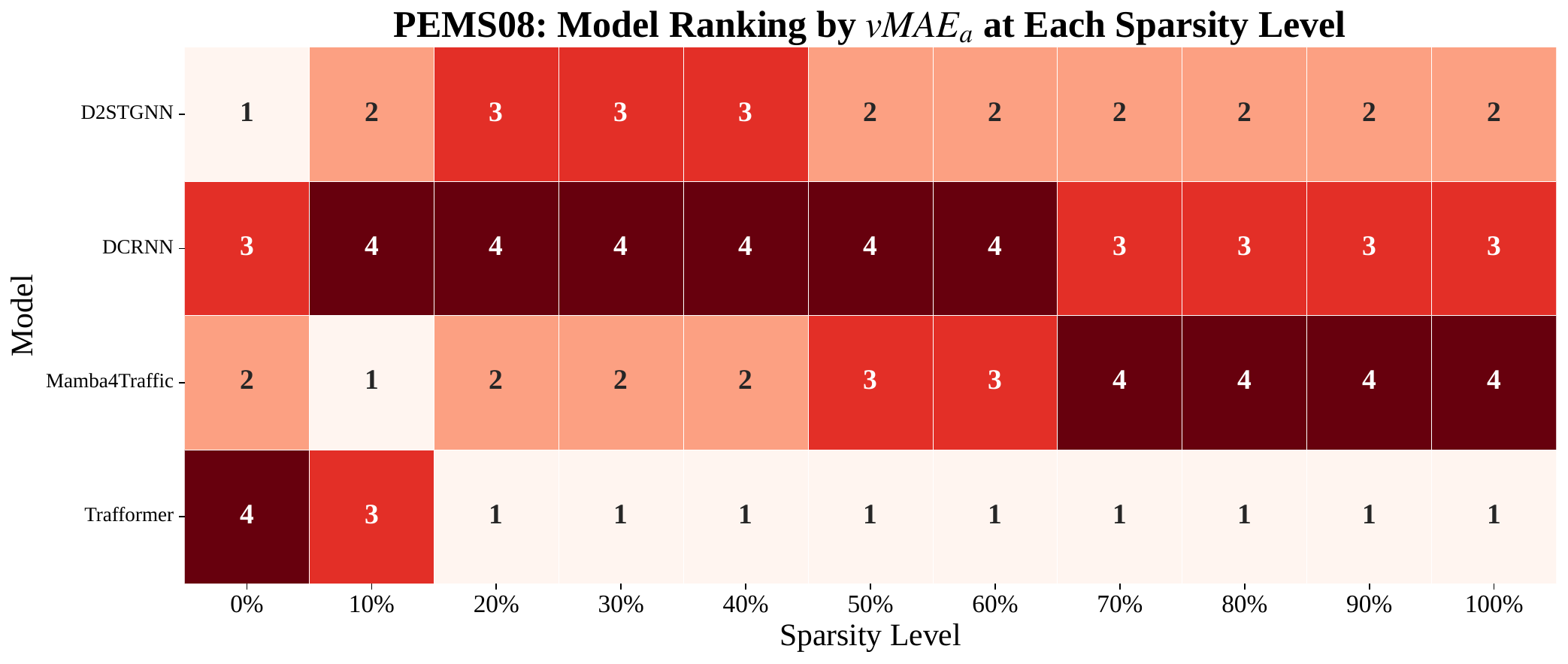}
    }
    \caption{Evolution of model efficacy rankings over the sparsity spectrum for (a) PEMS04 and (b) PEMS08. Cell values represent rank (1 = best vMAE); darker shading corresponds to poorer rank.}
    \label{fig:ranking_matrices}
\end{figure*}

Figures~\ref{fig:ranking_matrix_PEMS04} and~\ref{fig:ranking_matrix_PEMS08} summarize rank shifts across the complete grid $P_{10}$. D2STGNN ranks first only at $\rho = 0\%$. On both datasets, Mamba4Traffic is first at $10\%$. On PEMS04, Mamba4Traffic remains first through $50\%$ and Trafformer is first from $60\%$ through $100\%$. On PEMS08, Trafformer is first from $20\%$ through $100\%$. On PEMS04, D2STGNN is last from $10\%$ onward. On PEMS08, Mamba4Traffic falls to last place from $70\%$ onward, while D2STGNN returns to second. Full-data leadership and operational leadership therefore diverge as soon as $10\%$ of nodes are masked.

Figure~\ref{fig:accuracy_robustness_tradeoff} illustrates the relationship between baseline accuracy ($vMAE_a$ at zero sparsity) and robustness (weighted mean degradation $\bar{D}_{MAE}$ from Table~\ref{tab:sparsity_degradation}) for each model across the $0\%$ through $35\%$ sparsity range. On PEMS04, D2STGNN attains the best baseline ($vMAE_a=0.159$) but the highest weighted degradation ($4.84\times$), whereas Trafformer has the weakest baseline ($0.229$) and the lowest $\bar{D}$ ($1.78\times$). On PEMS08 the same pattern holds. D2STGNN leads at $\rho=0$ ($0.158$) with $\bar{D}=2.40\times$, and Trafformer has the weakest baseline ($0.259$) with the lowest $\bar{D}$ ($1.48\times$). Mamba4Traffic sits near D2STGNN on the accuracy axis and near DCRNN on the robustness axis. The dashed line shows the general direction of this tradeoff. Pearson $r$ is $-0.643$ on PEMS04 and $-0.659$ on PEMS08. Each model's baseline $vMAE_a$ at $\rho=0$ and its $\bar{D}_{MAE}$ are converted to standard scores over the four models, and $r$ is the covariance of those two standardized series. Negative $r$ means that better full-data accuracy is associated with larger weighted degradation. A model that is strongest at $\rho=0$ is therefore not also the most robust under sparsity, and a practical choice has to balance the two.

The same figure also reports a sensitivity check on the Gaussian weight. Circles are the $\alpha=20$ means. Black capped bars show $\pm 1$ sample standard deviation across mask seeds. Triangles and squares mark the same models at $\alpha=10$ and $\alpha=40$. Changing $\alpha$ shifts the degradation magnitudes but does not reverse which model is most or least robust. Trafformer remains first and D2STGNN remains last on both datasets at all tested values between $\alpha=10$ and $\alpha=40$. The accuracy versus robustness conclusion is therefore insensitive to reasonable variation in $\alpha$, although the reported magnitude of $\bar{D}_{MAE}$ remains weighting-dependent.

\begin{table*}[!t]
\caption{Channel-averaged vehicular MAE ($vMAE_a$) at 10\% sparsity for each model and horizon on PEMS04 and PEMS08. Values are means over three mask seeds (43, 44, and 45). Best and second-best per column are bold and underlined.}
\label{tab:10percent_sparsity}
% Auto-generated by scripts/generate_multiseed_tables_figures.py — do not edit by hand.
% Channel-averaged vMAE at 10% sparsity by horizon; mean over mask seeds 43--45.

\begin{tabular*}{\textwidth}{@{\extracolsep{\fill}}l l cccccccccccc@{}}
\hline
Dataset & Model & H1 & H2 & H3 & H4 & H5 & H6 & H7 & H8 & H9 & H10 & H11 & H12 \\
\hline
\noalign{\vskip 0.4em}
\multirow{4}{*}{PEMS04} & DCRNN          & \underline{0.407} & 0.409 & 0.408 & 0.407 & 0.407 & 0.406 & 0.407 & 0.407 & 0.408 & 0.409 & \underline{0.411} & \underline{0.413} \\
& D2STGNN        & 0.430 & 0.441 & 0.451 & 0.496 & 0.503 & 0.509 & 0.627 & 0.632 & 0.637 & 1.080 & 1.091 & 1.098 \\
& Mamba4Traffic  & \textbf{0.291} & \textbf{0.305} & \textbf{0.315} & \textbf{0.318} & \textbf{0.323} & \textbf{0.325} & \textbf{0.327} & \textbf{0.327} & \textbf{0.327} & \textbf{0.327} & \textbf{0.325} & \textbf{0.326} \\
& Trafformer     & 0.515 & \underline{0.405} & \underline{0.394} & \underline{0.395} & \underline{0.397} & \underline{0.399} & \underline{0.401} & \underline{0.395} & \underline{0.403} & \underline{0.405} & 0.411 & 0.423 \\
\noalign{\vskip 0.4em}
\hline
\noalign{\vskip 0.4em}
\multirow{4}{*}{PEMS08} & DCRNN          & 0.427 & 0.429 & 0.431 & 0.432 & 0.432 & 0.433 & 0.434 & 0.435 & 0.435 & 0.436 & 0.437 & 0.438 \\
& D2STGNN        & \underline{0.409} & 0.417 & 0.422 & 0.400 & 0.401 & 0.401 & \underline{0.381} & \underline{0.380} & \underline{0.380} & \underline{0.365} & \underline{0.365} & \underline{0.366} \\
& Mamba4Traffic  & \textbf{0.296} & \textbf{0.303} & \textbf{0.307} & \textbf{0.311} & \textbf{0.312} & \textbf{0.316} & \textbf{0.316} & \textbf{0.320} & \textbf{0.322} & \textbf{0.324} & \textbf{0.329} & \textbf{0.339} \\
& Trafformer     & 0.445 & \underline{0.398} & \underline{0.394} & \underline{0.391} & \underline{0.385} & \underline{0.384} & 0.390 & 0.390 & 0.390 & 0.390 & 0.384 & 0.384 \\
\noalign{\vskip 0.4em}
\hline
\end{tabular*}

\end{table*}

\begin{figure*}[!t]
    \centering
    \subfloat[PEMS04]{\label{fig:opbreak_PEMS04}
        \includegraphics[width=0.475\textwidth]{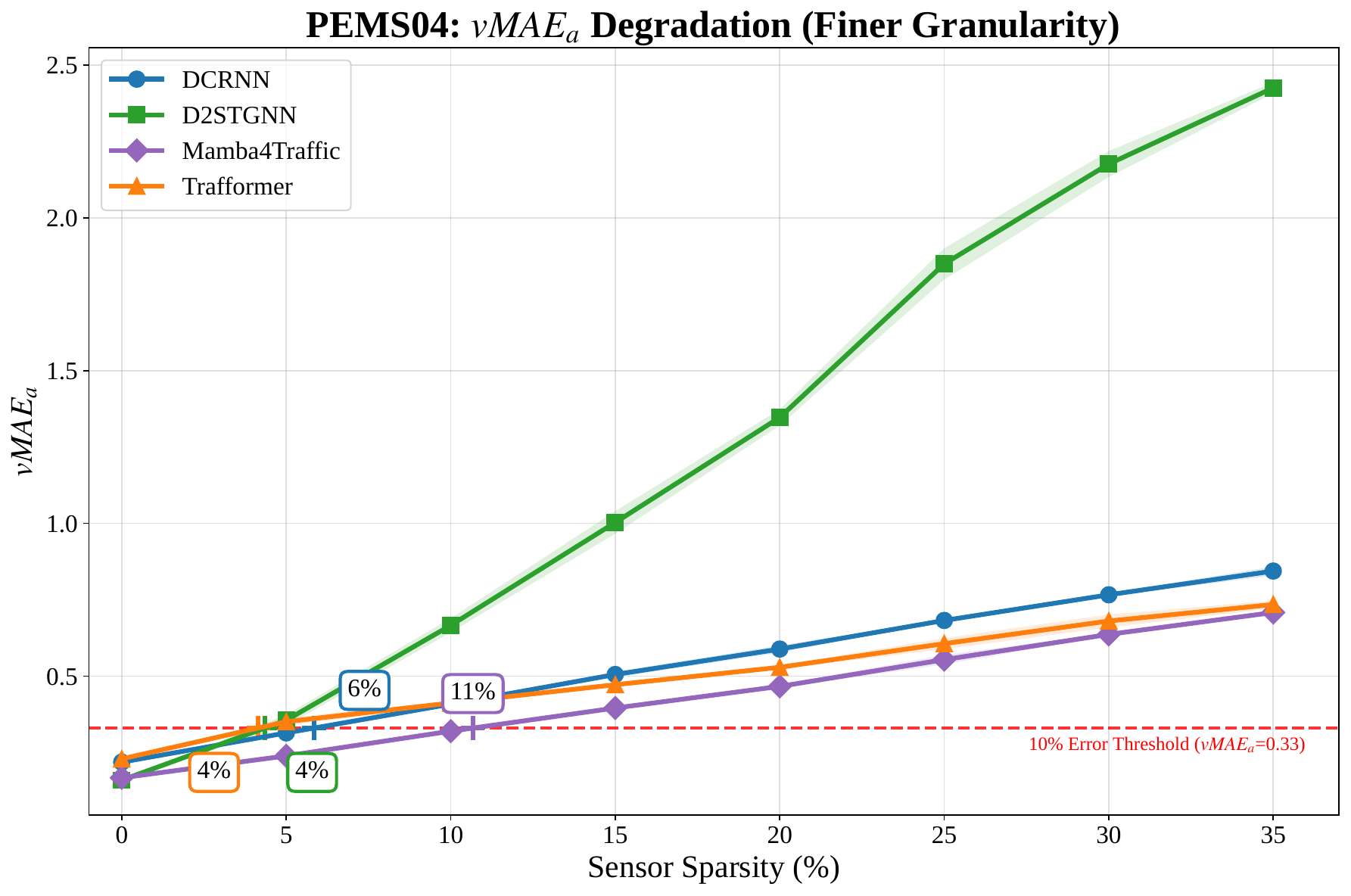}
    }\hfill
    \subfloat[PEMS08]{\label{fig:opbreak_PEMS08}
        \includegraphics[width=0.475\textwidth]{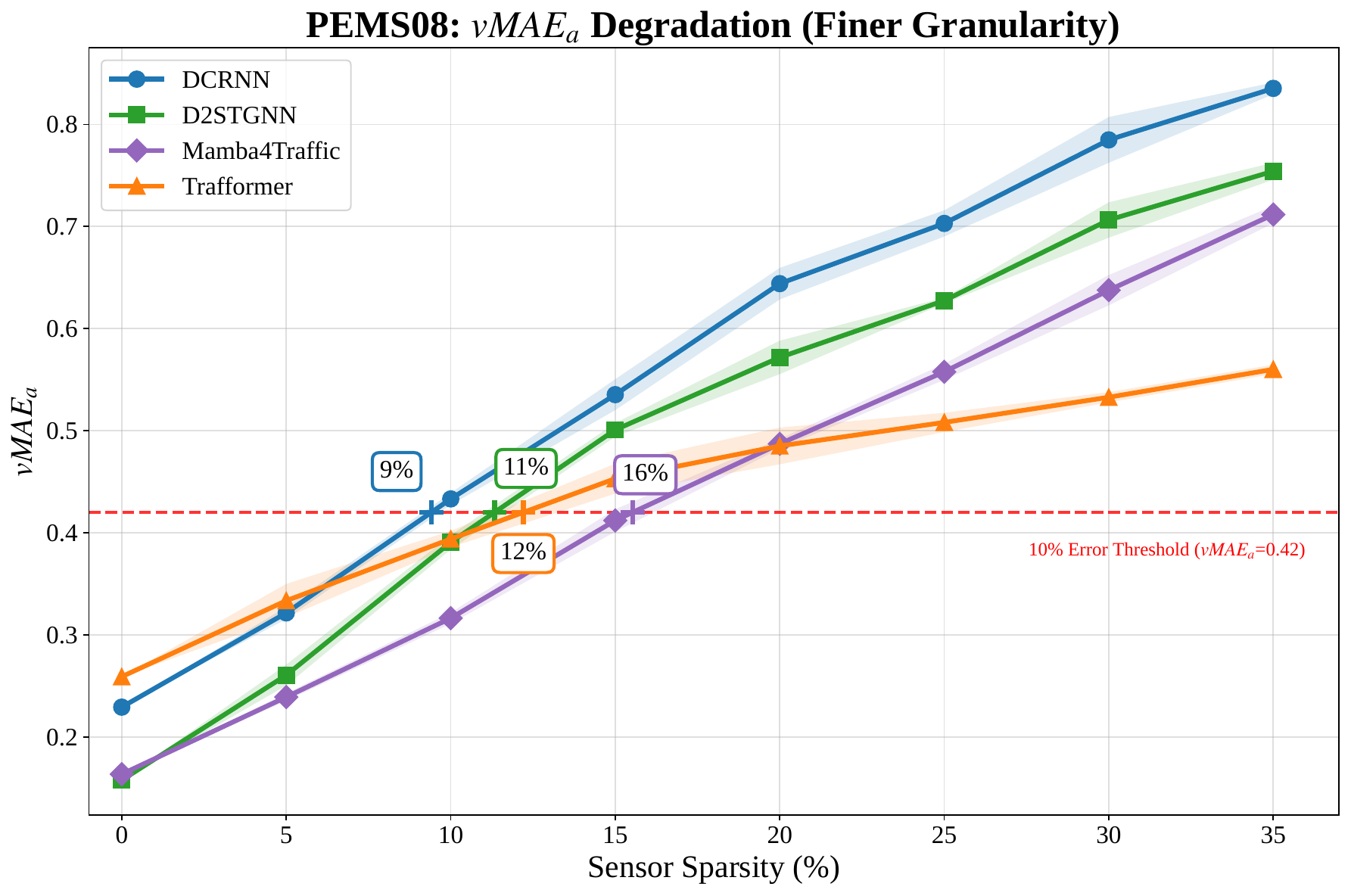}
    }
    \caption{Operational breakdown analysis showing $vMAE_a$ curves from 0--35\% sparsity at 5\% intervals for (a) PEMS04 ($\tau=0.33$) and (b) PEMS08 ($\tau=0.42$). Dashed lines indicate the dataset-specific 10\%-error usability threshold, with markers denoting interpolated breakdown points.}
    \label{fig:operational_breakdown}
\end{figure*}

\begin{figure*}[!t]
    \centering
    \subfloat[PEMS04]{\label{fig:perc_increase_PEMS04}
        \includegraphics[width=0.475\textwidth]{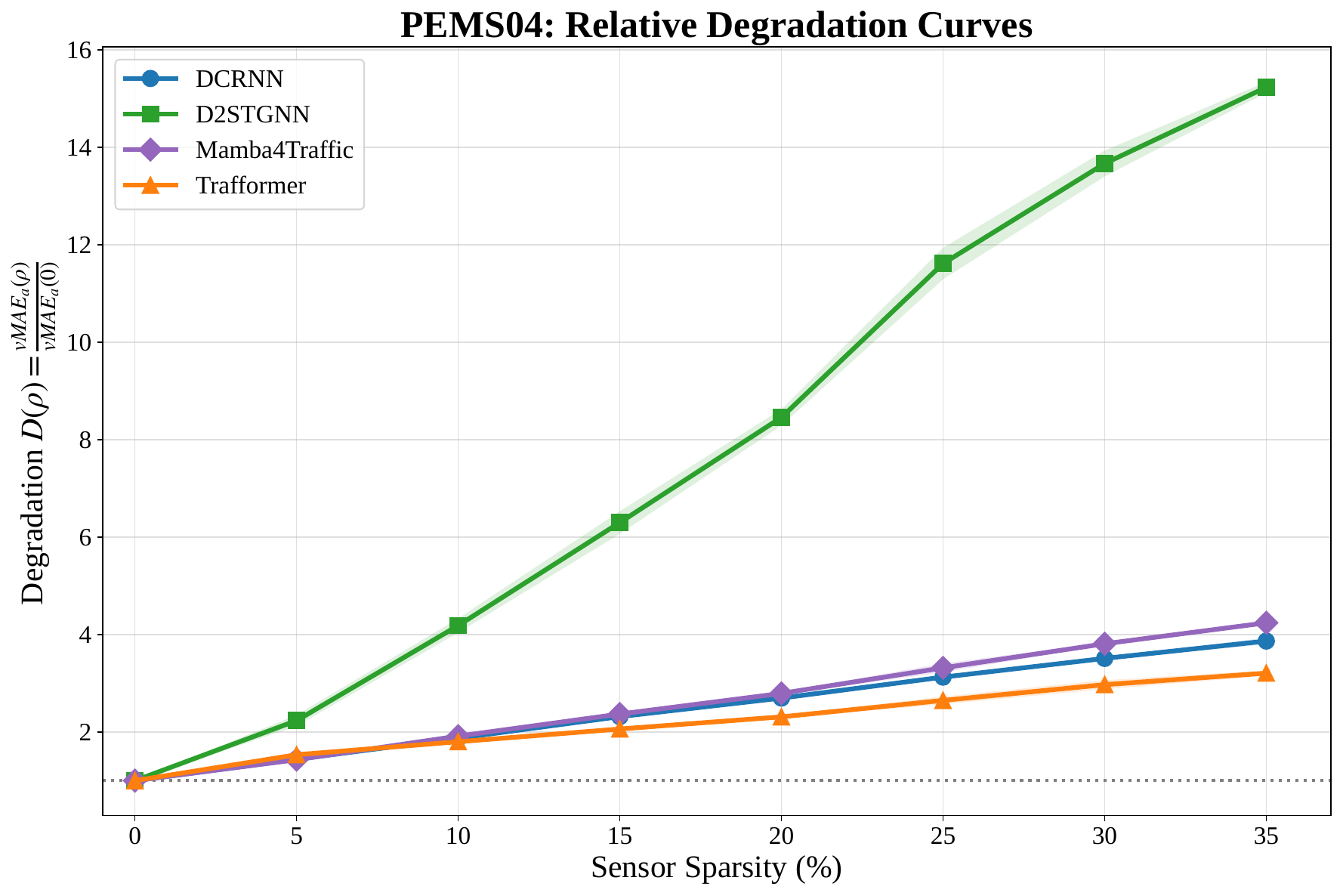}
    }
    \hfill
    \subfloat[PEMS08]{\label{fig:perc_increase_PEMS08}
        \includegraphics[width=0.475\textwidth]{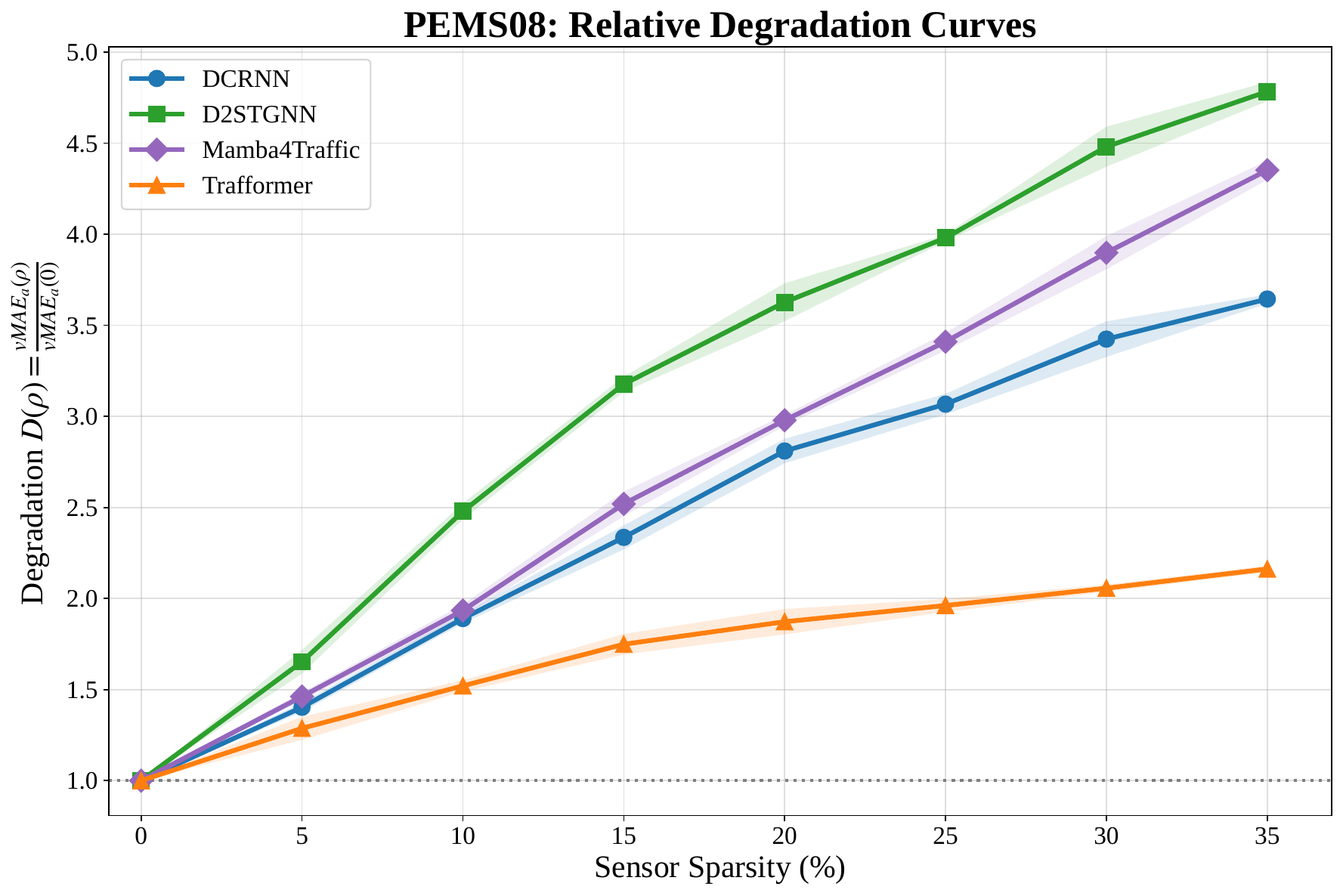}
    }
    \caption{Degradation $D(\rho)$ in vMAE as a function of input sparsity (0--35\% at 5\% intervals), relative to each model's baseline at 0\% sparsity, for (a) PEMS04 and (b) PEMS08.}
    \label{fig:percentage_increase_vmae}
\end{figure*}

\subsection{Operational Threshold Analysis}

The threshold analysis presented in Figures~\ref{fig:opbreak_PEMS04} and~\ref{fig:opbreak_PEMS08} quantifies the sparsity level, denoted as a percentage, at which each model's prediction error exceeds a $10\%$ deviation from the ground truth. Specifically, it identifies where the normalized $vMAE_a$ surpasses $0.33$ for PEMS04 or $0.42$ for PEMS08. This threshold offers an operationally meaningful measure of model robustness, signifying the point beyond which predictions are no longer reliable for practical traffic management.

On PEMS08 ($\tau = 0.42$), Mamba4Traffic remains below the threshold until approximately $16\%$ sparsity, the highest among the evaluated models. Trafformer crosses at approximately $12\%$, D2STGNN at approximately $11\%$, and DCRNN at approximately $9\%$.

On PEMS04 the usability threshold is stricter ($\tau = 0.33$). Mamba4Traffic remains usable the longest (approximately $11\%$), followed by DCRNN (approximately $6\%$), then D2STGNN and Trafformer (both approximately $4\%$). The lower breakdown percentages on PEMS04 reflect the stricter normalized threshold, which arises from higher traffic variability under Equation~\ref{eq:tau}, not from a uniformly less robust architecture.

Overall, these results highlight a trade-off between baseline accuracy and robustness to sensor sparsity. D2STGNN is the most accurate model at full data, yet it and Trafformer both leave the usable range first on PEMS04 (approximately $4\%$). Mamba4Traffic has the latest PEMS04 crossing. Trafformer has the weakest baselines and the lowest weighted $\bar{D}$ on both datasets.

The relative degradation curves shown in Figures~\ref{fig:perc_increase_PEMS04} and~\ref{fig:perc_increase_PEMS08} decouple sparsity robustness from baseline accuracy over the 0--35\% operational grid. D2STGNN shows the steepest relative error growth on PEMS04. Mamba4Traffic is next on both datasets. Trafformer stays closest to its baseline through 35\% sparsity. DCRNN lies between Trafformer and Mamba4Traffic and is clearly more stable than D2STGNN.

\subsection{Per-horizon analysis}

The horizon-wise analysis of $vMAE_a$ at 10\% sparsity is summarized in Table~\ref{tab:10percent_sparsity}. At $10\%$ sparsity, Mamba4Traffic has the lowest $vMAE_a$ on every horizon of both datasets. The full-data ordering therefore does not carry over unchanged once $10\%$ of sensors are masked, and the ranking is not an artifact of a single prediction step.

\section{Discussion}
\label{sec:discussion}

The results show a persistent gap between full-data accuracy and masking tolerance. D2STGNN is the most accurate model at $\rho = 0$ on both datasets, yet it has the largest weighted degradation, especially on PEMS04. Mamba4Traffic is a close second at full data and remains competitive at $10\%$ sparsity, but it is not the robustness leader. Trafformer has weaker baselines and the lowest $\bar{D}$. It has the best absolute error from $\rho = 60\%$ on PEMS04 and from $\rho = 20\%$ on PEMS08. DCRNN is in the middle of the $\bar{D}$ ranking. No single model is best on baseline accuracy, weighted degradation, and breakdown point together.

Although usability thresholds differ between PEMS04 and PEMS08 ($\tau = 0.33$ versus $0.42$), that difference comes from dataset variability, not from a change in which models are usable. Models leave the usable range in a similar operational band, about $4\%$--$11\%$ sparsity on PEMS04 and $9\%$--$16\%$ on PEMS08. The breakdown point and $\bar{D}$ therefore answer different questions. One says when a model becomes unusable. The other says how fast error grows relative to that model's own full-data performance.

Finally, the introduction of the usability threshold serves as a practical decision-making tool for transportation engineers. By quantifying the breakdown point as a specific sparsity percentage, we move away from abstract error values toward a safety rating for models. This allows practitioners to select models based on the known reliability of their sensing infrastructure. For instance, if a sensor network is expected to lose about $10\%$ of detectors, selecting a model whose breakdown point lies above that rate is a more rational engineering choice than selecting a model with higher peak accuracy but an earlier crossing.

Further study can examine spatially correlated and temporally correlated sensor failures, as well as alternative missing-input representations such as last-observation carry-forward and mean fill.

\section{Conclusion}
\label{sec:conclusion}

In this paper, we presented a comprehensive sparsity analysis of multivariate traffic prediction models, addressing a critical gap in the literature regarding the evaluation of model robustness under inference-time sensor data loss. While most research focuses on baseline accuracy under ideal conditions, our work systematically quantified the performance degradation of diverse architectural paradigms across the full spectrum of sensor sparsity. By employing normalized vehicular metrics and introducing the concept of the usability breakdown point, we provided a framework for assessing the operational reliability of models in real-world scenarios.

The summary of our findings reveals a fundamental accuracy-robustness trade-off that should guide future model selection. We observed that models with the highest baseline accuracy, such as D2STGNN, are often the most sensitive to input sparsity, with performance degrading by up to $4.84\times$ relative to their full-data baseline within the operationally relevant 0--35\% sparsity range. Conversely, models with lower peak performance, such as Trafformer, can exhibit more stable relative behavior ($1.78\times$ / $1.48\times$) and take the lead in absolute error at high sparsity, from $60\%$ missing sensors on PEMS04 and from $20\%$ on PEMS08. These results emphasize that the model with the best accuracy under ideal conditions may be the least reliable for deployment in real-world scenarios where sensor data is frequently missing.

We do not identify a single architecture as universally optimal among the models evaluated. Model choice should depend on the deployment constraint that matters most. D2STGNN is preferable when full-data accuracy is the only objective. Mamba4Traffic is preferable when 10--20\% sparsity is expected. It keeps a strong full-data baseline, leads absolute error on PEMS04 through $50\%$, and has the latest PEMS04 breakdown. Trafformer is preferable when the criterion is weighted relative degradation, or when sparsity is severe, because it keeps the lowest $\bar{D}$ and leads absolute error from $60\%$ on PEMS04 and from $20\%$ on PEMS08. DCRNN remains a middle option on $\bar{D}$ but is not the breakdown leader on either dataset. Deployment decisions should therefore account for expected sensor loss, dataset-specific baseline calibration, and which metric defines operational success.

Ultimately, for traffic forecasting to successfully transition from controlled research environments to the high-stakes environment of city control rooms, robustness must be treated as a primary performance metric rather than an afterthought. Our results demonstrate that peak accuracy is an insufficient proxy for real-world suitability. Future research should prioritize the development of sparsity-aware training paradigms, where random masking is integrated directly into the training phase. By bridging the gap between baseline excellence and architectural resilience, we can move toward a new generation of traffic models that remain reliable under the imperfect and dynamic conditions inherent to modern urban infrastructure.

Our code, pretrained checkpoints, and evaluation scripts are publicly available at \url{https://github.com/mombash/Libcity-based-implementations}.

\FloatBarrier

\bibliographystyle{IEEEtran}
\bibliography{references}
\end{document}